\documentclass[11pt]{article}

\usepackage[final]{acl}

\usepackage{times}
\usepackage{latexsym}

\usepackage[T1]{fontenc}
\usepackage[utf8]{inputenc}

\usepackage{microtype}

\usepackage{inconsolata}

\usepackage{graphicx}

\usepackage{amsmath}
\usepackage{amssymb}
\usepackage{array}
\usepackage{xcolor}      % For colored cells
\usepackage{colortbl}    % For table coloring
\usepackage{multirow}    % For merged cells
\usepackage{booktabs}
\usepackage{pifont}
\usepackage{makecell}
\usepackage{svg}
\usepackage{siunitx}
\usepackage{arydshln}
\usepackage{threeparttable}
\usepackage{wrapfig}
\usepackage[table]{xcolor}

\usepackage{amsmath,amsfonts,bm}

\def\eqref#1{equation~\ref{#1}}
\def\1{\bm{1}}

\DeclareMathAlphabet{\mathsfit}{\encodingdefault}{\sfdefault}{m}{sl}
\SetMathAlphabet{\mathsfit}{bold}{\encodingdefault}{\sfdefault}{bx}{n}

\usepackage{hyperref}
\usepackage{url}

\usepackage{morewrites}
\usepackage{bbm}
\usepackage{bm}
\usepackage{color}
\usepackage{multirow}
\usepackage{array}
\usepackage{booktabs}
\usepackage{diagbox}
\usepackage{bbding}
\usepackage{colortbl}
\usepackage{tabularx}
\usepackage{makecell}
\usepackage[dvipsnames]{xcolor}
\usepackage{hyperref}
\usepackage{subfigure}
\usepackage{amsmath, amssymb, overpic, textpos}
\usepackage{algorithmicx}
\usepackage{algpseudocode}
\usepackage{listings}
\usepackage{multicol}
\usepackage{xspace}
\usepackage{wrapfig}
\usepackage{graphicx}
\usepackage{graphbox}
\usepackage{arydshln}
\usepackage{algorithm}
\usepackage{adjustbox}
\usepackage{kotex}
\usepackage{caption}
\usepackage{setspace} 
\usepackage{epigraph}
\usepackage[most]{tcolorbox}
\usepackage{cleveref}

\definecolor{uclablue}{rgb}{0.15, 0.45, 0.68}
\hypersetup{
    breaklinks,
    citecolor=uclablue,
    colorlinks=true,
}

\definecolor{mygray}{gray}{0.6}

\usepackage{cleveref}
\usepackage{bbm}
\usepackage{wrapfig}

\definecolor{my_green}{RGB}{51,102,0}
\definecolor{my_red}{RGB}{204, 0, 0}
\definecolor{my_purple}{RGB}{160, 43, 147}
\definecolor{my_blue}{RGB}{15, 158, 213}
\definecolor{darkgrey}{rgb}{0.53,0.53,0.53}
\definecolor{mygrey}{rgb}{0.9,0.9,0.9}
\definecolor{circlered}{rgb}{252,100,84}

\definecolor{th_green}{rgb}{103,159,108}
\definecolor{code}{rgb}{59,94,176}
\definecolor{answer}{rgb}{182,27,25}
\definecolor{interp}{rgb}{0.53,0.53,0.53}

\newcommand{\bench}{\textsc{OceanGym}}
\definecolor{green}{RGB}{51,102,0}
\definecolor{red}{RGB}{204, 0, 0}
\definecolor{purple}{RGB}{160, 43, 147}

\title{OceanGym: Evaluating Language-Grounded Embodied Agents in Underwater Environments}

\author{
    Yida Xue$^{\spadesuit\diamondsuit}$,
    Mingjun Mao$^{\clubsuit\diamondsuit}$,
    Xiangyuan Ru$^{\spadesuit\diamondsuit}$,
    Yuqi Zhu$^{\spadesuit\diamondsuit}$,
    Baochang Ren$^{\spadesuit\diamondsuit}$,
    Shuofei Qiao$^{\spadesuit\diamondsuit}$,\\
    \textbf{Mengru Wang$^{\spadesuit\diamondsuit}$},
    \textbf{ Shumin Deng$^{\spadesuit\diamondsuit}$,
    Xinyu An$^{\spadesuit\diamondsuit}$,
    Ningyu Zhang$^{\spadesuit\diamondsuit}$\thanks{$\quad$ Corresponding Author.},
    Ying Chen$^{\spadesuit\diamondsuit}$,
    Huajun Chen$^{\spadesuit\diamondsuit}$}\\
    $^\spadesuit$Zhejiang University ~
    $^\clubsuit$National University of Singapore\\
    $^\diamondsuit$State Key Laboratory of Ocean Sensing, Zhejiang University\\
    \fontsize{10.2pt}{0.1\baselineskip}\selectfont \texttt{\{xueyida,zhangningyu,huajunsir\}@zju.edu.cn}\\
    \raisebox{-1.2pt}{\includegraphics[scale=0.03]{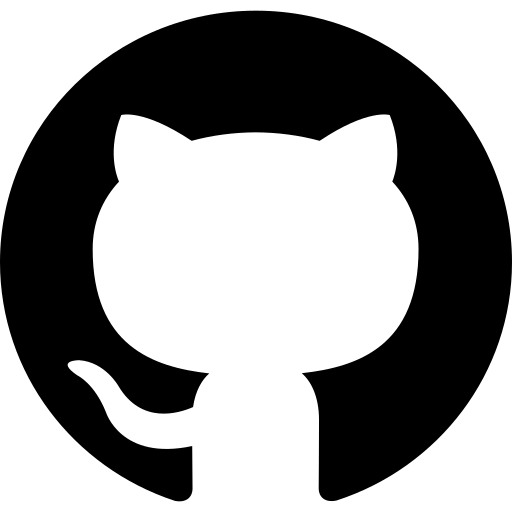}}\,\url{https://oceangpt.github.io/OceanGym}
}

\begin{document}
\maketitle

\begin{abstract}
We introduce {\bench}, the first comprehensive benchmark for underwater embodied agents, designed to advance AI in one of the most demanding real-world environments. {\bench} features eight realistic task domains and a unified agent framework driven by Multimodal Large Language Models (MLLMs) that integrates perception, memory, and sequential decision-making. Agents must follow natural language instructions, interpret optical and sonar inputs, autonomously explore complex environments, and achieve long-horizon goals. Extensive experiments reveal a significant performance gap between state-of-the-art MLLM-driven agents and human experts, underscoring persistent challenges in perception, planning, and adaptation underwater. By offering a high-fidelity, rigorously designed platform, {\bench} establishes a testbed for robust embodied AI and represents a foundational step toward sim-to-real transfer on Autonomous Underwater Vehicles (AUVs), marking progress toward agents capable of operating in one of Earth's last unexplored 
frontiers\footnote{The code and data are available at \url{https://github.com/OceanGPT/OceanGym}.}.
\end{abstract}

\section{Introduction}

\begin{figure*}[!ht]
\centering
\includegraphics[width=0.99\textwidth]{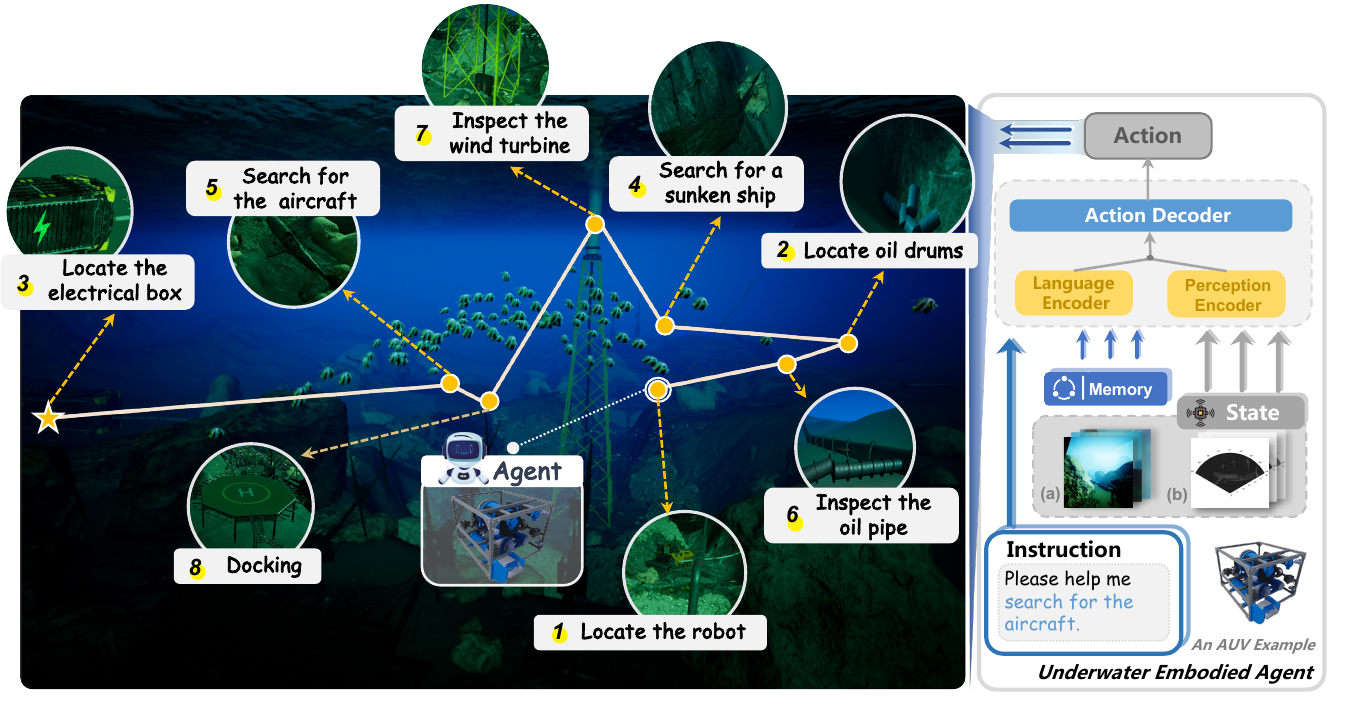}
\caption{
{\bench} introduces a unified \textbf{agent framework} across \textbf{8 real-world underwater scenarios}.
The agent interprets instruction, utilizes memory, fuses optical and sonar imagery, and controls Autonomous Underwater Vehicles (AUVs).
}
\label{fig:intro}
\end{figure*}

As Richard S. Sutton famously noted, we are entering an ``era of experience'' \citep{silver2025welcome}. 
Embodied agents equipped with language models \citep{zhao2023survey,gpt4o} are emerging as a central paradigm for intelligent systems, as they accumulate and leverage experience through continuous interaction to close the perception-decision-action loop in physical or simulated environments \citep{embodiedai1,embodiedai2,liu2025embodied}. Unlike static decision or generative models, these agents must integrate rich multimodal sensory streams and execute continuous-control policies to achieve long-horizon objectives. 
This necessitates a unified treatment of perceptual representation, planning, online inference, and sequential policy optimization \citep{fung2025embodied}. Significant progress has been demonstrated across diverse
domains \citep{r2r,ma2025position,EmbodiedCity}. 

In contrast, underwater embodied agents remain largely unexplored despite their critical scientific and engineering importance \citep{visbeck2018ocean,kelly2022empowering,zheng2023marinegpt,li2024ocean,gao2025neuralom}. 
Deploying embodied agents in marine environments offers unique opportunities for ocean exploration, offshore resource development, environmental monitoring, and subsea rescue operations. 
These tasks impose stringent requirements on the robustness and reliability of autonomous platforms, making the development of agents capable of functioning under real marine conditions a key bridge between simulated research and practical deployment \citep{ma2025state}.

\textbf{Challenges.}
Underwater embodied agents face distinct challenges that set them apart from overland and aerial systems. 
\emph{Perceptually}, poor visibility and low-light conditions, combined with the inherent limitations of optical sensors, compel reliance on sonar, inertial measurements, and other sparse modalities \citep{li2024dual,aubard2025sonar}. 
These heterogeneous and noisy observations complicate sensor fusion and perception. 
\emph{Environmentally}, deep-sea and offshore settings are largely unexplored, with unstable localization, absent prior knowledge, and dynamic currents. 
The lack of prior knowledge prevents effective environmental modeling, requiring agents to reason under circumstances of extreme partial observability and uncertainty \citep{sariman2025ur}. 
Together, these factors constrain the development of underwater agents, leaving their capabilities in early stages.

\textbf{Building OceanGym.} 
To address these challenges, we introduce {\bench}, an open environment benchmark for underwater embodied agents. 
{\bench} constructs a comprehensive marine environment spanning approximately 800m × 800m (length × width), with dynamically adjustable depth to simulate varying lighting conditions. The platform incorporates eight distinct task domains designed to reflect real-world operational requirements. Additionally, it provides a multimodal agent framework that integrates perception, memory, and action decision-making capabilities for controlling AUVs.
The benchmark unifies perception and decision-making in simulated underwater scenarios, where agents must infer target states from contextual cues or multi-view sensor data and execute complex behaviors such as search, inspection, and docking. 
By simulating these environments, {\bench} enables systematic evaluation of models’ capabilities in underwater embodied settings and offers an initial pathway for transferring learned skills to real-world AUVs.

\textbf{Benchmark Results and Insights.}
Extensive experiments demonstrate performance gaps between MLLMs and human experts, especially under low-visibility conditions. Agents often exhibit difficulties in accurately interpreting sonar data, distinguishing objects in complex scenes, and maintaining consistent decision-making strategies over prolonged missions. Additional limitations emerge in memory retention and adaptability when objects are occluded or environmental conditions change.

\section{OceanGym}
\begin{figure*}[t]
\centering
\includegraphics[width=0.99\textwidth]{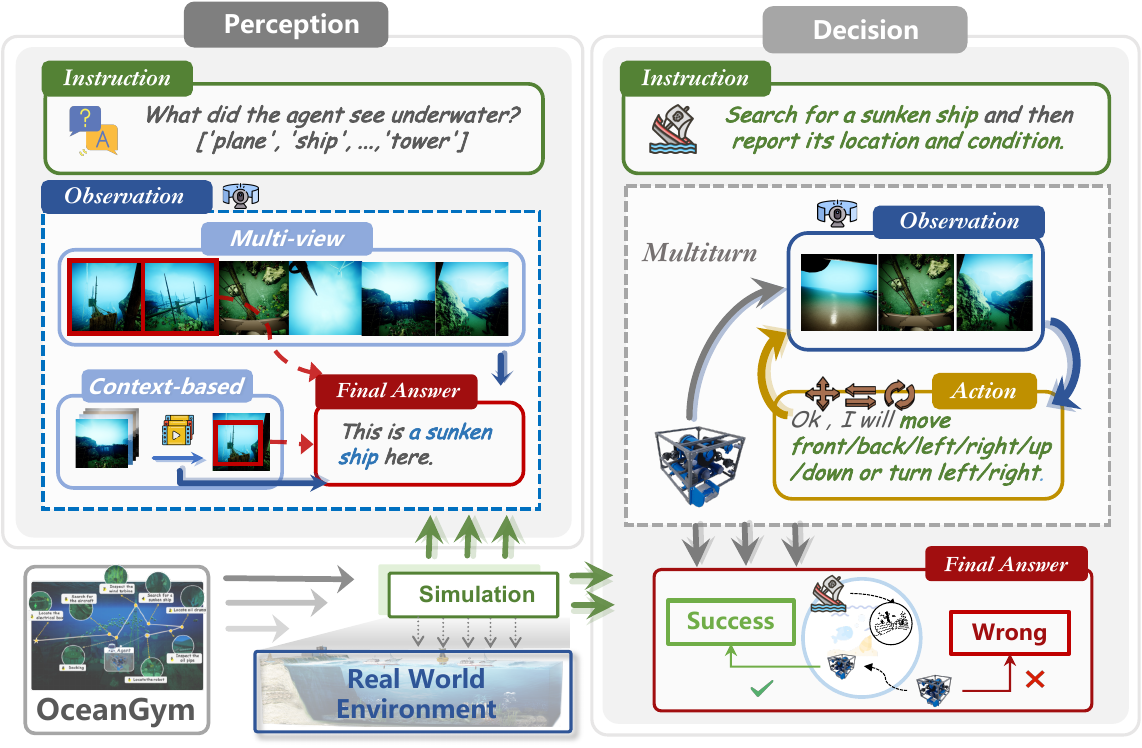}
\caption{
\textbf{{\bench} Tasks.} {\bench} comprises \textbf{Perception Tasks} (divided into \textbf{Multi-view Perception} and \textbf{Context-based Perception} settings) and \textbf{Decision Tasks} for evaluating embodied agents.
}
\label{fig:oceangym}
\end{figure*}

{\bench} is a high-fidelity embodied underwater environment that simulates a realistic ocean setting with diverse scenes. As illustrated in Figure~\ref{fig:oceangym}, {\bench} establishes a robust benchmark for evaluating autonomous agents through a series of challenging tasks, encompassing various perception analysis and decision-making tasks in navigation. {\bench} facilitates these evaluations by equipping MLLM-driven agents with multi-modal perception and parameterized action spaces.

% \begin{figure*}[t]
% \vspace{-0.8in}
% \centering
% \includegraphics[width=0.92\textwidth]{figure/pic2.pdf}
% \caption{\small
% \textbf{{\bench} Tasks.} {\bench} comprises \textbf{Perception Tasks} (divided into \textbf{Multi-view Perception} and \textbf{Context-based Perception} settings) and \textbf{Decision Tasks} for evaluating embodied agents.
% }
% \label{fig:oceangym}
% \vspace{-0.5cm}
% \end{figure*}

\subsection{OceanGym Environment}
\label{sec:env}
We develop {\bench} atop Unreal Engine (UE) 5.3\citep{unrealengine}, which provides a comprehensive set of underwater environments, including natural terrains and engineered structures.
The environment features several semantic regions such as open water, seabed plains, underwater cliffs, pipeline networks, wreckage sites, and energy infrastructure zones. 
Each region is modeled with realistic physical and geometric properties, incorporating elements like oil pipelines, chemical waste barrels, submerged shipwrecks, electrical equipment, wind turbine foundations, and aircraft debris (more details in Appendix \ref{app:navigation}). 
% These elements are built using intricate 3D assets based on real-world references, ensuring accurate representation of structural and material characteristics. 
We simulate varying lighting conditions by adjusting underwater depth, considering two representative scenarios: shallow (50m) and deep (500m) water. In deep water, optical sensing depends on artificial lighting with a visibility range of about 0–10m.
Moreover, {\bench} is scalable: users can customize the environment by adjusting depths for more complex lighting, introducing new objects, or designing additional tasks within the existing framework, thereby extending diversity and difficulty.

\subsection{Underwater Embodied Agents}
We model the agent's control–perception loop as a Partially Observable Markov Decision Process (POMDP) enhanced with contextual memory.
At each time step \( t \), the agent processes the task specification \(\mathcal{T}=(I_{\text{target}},c)\), where \( I_{\text{target}} \) is a visual reference image of the target and \( c \) provides its textual identity and features. It also considers \textbf{language instruction \( L \), synchronized observations \( O_t \), and its memory state \( m_t \).} These elements collectively shape the agent's perception and objectives.

With the above information, the agent must generate either a textual perception response \( y_t \) for perception tasks, or determine a control action \( a_t \) for decision tasks. Here, \( a_t \in \mathcal{A} \) is a discrete action selected from the action space  \( \mathcal{A} \). A decision trajectory is described by $ \sigma =(O_1, a_1, s_1, m_1, \ldots, m_{t-1}, O_t, a_t, s_t)$. In this sequence, \( O_i \) represents the observations, \( a_i \) the actions, \( s_i \) the states, and \( m_i \) the memory states at each time step \( i \). The episode concludes when the target is achieved or when the maximum time \( t_\textrm{max} \) is reached. The ultimate reward is based on the successful score of the task, as defined in \S \ref{metrics}.

\textbf{State and Observation.} 
The agent's state at time $t$ is given by $s_t=\{(x_t, y_t, z_t), (\phi_t, \theta_t, \psi_t)\}$, where $(x_t, y_t, z_t)$ represent the agent's 3D positional coordinates, and $(\phi_t, \theta_t, \psi_t)$ denote the roll, pitch, and yaw angles, respectively.
At each timestep, the agent receives synchronized RGB and sonar images from sensors oriented in six different directions. The directions are defined by the set $\mathcal{D}_\text{sensor}=\{\text{f}, \text{b}, \text{l}, \text{r}, \text{u}, \text{d}\}$, corresponding to \texttt{front}, \texttt{back}, \texttt{left}, \texttt{right}, \texttt{up}, and \texttt{down}.
The RGB images from these directions are denoted as $O_t^R=\{o^R_{t,d}\}_{d \in \mathcal{D}_\text{sensor}}$, and the sonar images are represented similarly as $O_t^S=\{o^S_{t,d}\}_{d \in \mathcal{D}_\text{sensor}}$. The complete observation at time $t$ can be expressed as a combination of both image sets, $O_t=(O_t^R, O_t^S)$.

\textbf{Action Space.}
The agent's action direction set is defined as \(\mathcal{D}_\text{action} = \{f, b, l, r, u, d, rl, rr\}\), which encompasses both directional and rotational movements. Directional actions include translations along the primary axes: \texttt{forward} (\(f\)), \texttt{backward} (\(b\)), \texttt{left} (\(l\)), \texttt{right} (\(r\)), \texttt{up} (\(u\)), and \texttt{down} (\(d\)). Rotational actions consist of \texttt{rotate left} (\(\text{rl}\)) and \texttt{rotate right} (\(\text{rr}\)). At each timestep \(t\), the agent selects an action \(a_t \in \mathcal{A}\) from this discrete set and applies a control magnitude \(\delta \in \mathbb{R}_{\geq 0}\) to determine the execution intensity.

\textbf{Memory.}
Memory systems play a crucial role in storing and structuring past information, thereby enhancing the agent's resilience in dynamic and partially observable environments \citep{agent_survey, mem2, mem3, mem4, mem5}.
{\bench} agent maintains an explicit memory $m_t$, structured as a sliding window that records the last $K$ steps:
\begin{align}
m_t = \{(d_{t-k}, a_{t-k}) \mid k = 1, 2, \ldots, K\}
\end{align}
Within this memory structure, $d_{t-k}$ denotes the textual description at time $t-k$, and $a_{t-k}$ represents the corresponding action executed. The sliding window size $K$ is implemented primarily to prevent the context length from exceeding the model's maximum input capacity. The default window size is large enough to capture the necessary historical information for most tasks in our benchmark. The perception module $\mathcal{P}_\theta$, modeled as an MLLM, generates a summary based on the current context and the interaction history $\{(O_k, a_k)\}_{k=t-K}^{t}$:
\begin{align}
d_t = \mathcal{P}_\theta\!\left( \{(O_k,a_k)\}_{k=t-K}^{t}\right)
\end{align}
This summary is subsequently used to refresh the memory: $m_{t+1} = \mathrm{update}(m_t, d_t, a_t)$.

\textbf{Memory-augmented Markov Process.}
To maintain the Markov property while incorporating memory, we introduce an augmented state \(\tilde{s}_t = (s_t, m_t)\). The state transition is modeled as:
\begin{align}
p(\tilde{s}_{t+1} \mid \tilde{s}_t, a_t, \delta)
\end{align}
where \(p(\cdot \mid \cdot)\) represents the augmented state transition function of the environment. This function captures the evolution of memory, ensuring that the system remains Markovian despite the added complexity of memory integration.

\textbf{Agent Policy.}
The agent policy is a multimodal, memory-augmented mapping parameterized by an MLLM with parameter vector $\theta$:
\begin{align}
\pi_\theta(a_t,y_t \mid L,O_t,m_t,\mathcal{T},\delta)
\end{align}
Concretely, for perception tasks, we sample an answer $y_t \sim \pi_\theta(y \mid L,O_t,m_t,\mathcal{T},\delta)$, 
and for decision tasks, we sample an action $a_t \sim \pi_\theta(a \mid L,O_t,m_t,\mathcal{T},\delta)$.
An episode terminates at time $T$ when the agent either outputs a \textsc{Stop} command (for decision tasks) or provides a final answer to the question (for perception tasks) or when the maximum time $t_{\max}$ is reached. The policy, combined with the memory-augmented transition dynamics, induces the trajectory distribution:
\begin{align}
\begin{split}
\mathbb{P}_\theta(\sigma \mid L, \mathcal{T}) 
&= \prod_{t=1}^{T-1} \pi_\theta(a_t, y_t \mid L, O_t, m_t, \mathcal{T}, \delta) \\
&\quad \times p(\tilde{s}_{t+1} \mid \tilde{s}_t, a_t, \delta)
\end{split}
\end{align}
where \(\sigma\) represents the trajectory of the agent through the state space over time, influenced by the specified policy \(\pi_\theta\) and the transition model \(p(\tilde{s}_{t+1} \mid \tilde{s}_t, a_t, \delta)\).

\begin{tcolorbox}[
    colback=gray!5,
    colframe=blue!50!black,
    title=\textbf{Running Example: Shipwreck Area},
    fonttitle=\bfseries,
    boxrule=0.5pt,
    arc=1pt,
    left=4pt, right=4pt, top=4pt, bottom=4pt
]

\textbf{Perception Task:}
(1) Multi-view perception setting. 
The agent receives perception images (visual and sonar) from different sensors at the same time to determine the target, such as whether it is a shipwreck.
(2) Context-based perception setting.
The agent analyzes images one by one along a trajectory from a fixed viewpoint to identify the target.

\vspace{6pt}
\hrule
\vspace{6pt}

\textbf{Decision Task:}
The agent receives a task instruction, such as ``Search for a sunken ship,'' and then explores for 30 minutes to complete it.

\end{tcolorbox}

\subsection{OceanGym Perception Tasks}
\label{sec:task1}

The perception tasks are categorized into two settings: \textbf{Multi-View Perception} and \textbf{Context-based Perception}. These tasks primarily use RGB images as input, with sonar data added in certain experiments to enhance perception. The data for each setting are collected by human operators and designed to evaluate different aspects of MLLMs' perceptual abilities. There are a total of 85 scenes, and we also provide evaluation results on additional scenes. More details in Appendix~\ref{app:perception}.

\textbf{Multi-view Perception Setting.}
This setting evaluates the agent's ability to interpret visual information from multiple synchronized viewpoints. At each timestep $t$, the agent captures a set of six simultaneous RGB images, denoted as $O^{R}_{t} = \{o^{R}_{t,d}\}_{d \in \mathcal{D}_\text{sensor}}$, where $d$ refers to the different sensor orientations: \texttt{front}, \texttt{back}, \texttt{left}, \texttt{right}, \texttt{up}, and \texttt{down}. The objective is to consistently identify and localize underwater objects across these varied viewpoints. 
This setting examines whether objects visible from certain angles can be correctly perceived when the visual inputs from all directions are sequentially processed by the MLLM, thereby evaluating robustness to viewpoint variations.

\textbf{Context-based Perception Setting.}
This setting assesses the agent's ability to perceive and interpret sequential observations gathered during navigation. At each timestep $t$, the agent captures an RGB image $o^{R}_{t}$ from a fixed orientation, forming a chronological sequence $O^{R}_{1:m} = \{o^{R}_{t}\}_{t=1}^m$, where $m$ is the total number of timesteps. The agent must track and understand changes over time, ensuring consistent and accurate identification and localization of underwater objects. This evaluation emphasizes temporal consistency and the agent's capacity to build a coherent perception from a stable visual perspective in dynamic underwater environments.

\subsection{OceanGym Decision Tasks}
\label{sec:task2}
\paragraph{Decision Task Definition.}
Decision tasks evaluate decision-making in continuous 3D environments, where agents must integrate multimodal sensory input with task specifications. Each episode begins from an initial state $s_{0}=\{(x_{0},y_{0},z_{0}),(\phi_{0},\theta_{0},\psi_{0})\}$ and requires the agent to reach the target defined by $\mathcal{T}$. The agent must combine sensory observations $O_{t}$, temporal memory, and goal information to execute precise maneuvers in cluttered, low-visibility environments. Key parameters of the task include the decision interval $t_{\text{interval}}$ and the task's limited duration $t_{\max}$\footnote{By default, $t_{\text{interval}}$ takes 30 seconds and $t_{\max}$ takes 0.5 hours in decision tasks.}. The decision interval $t_{\text{interval}}$ determines how frequently the agent makes decisions and executes actions. The total task duration $t_{\max}$ sets the temporal constraint, within which the agent must meet its objectives, thereby influencing the planning and movement strategies employed by the agent. Compared with grid-based navigation benchmarks, this task emphasizes continuous control and realistic underwater environment, reflecting the challenges of autonomous exploration and inspection tasks.

\paragraph{Decision Task Design.}
To evaluate the decision-making capabilities of MLLMs in marine environments, we design eight representative task scenarios that are commonly used in actual underwater operations (more details in Appendix~\ref{app:navigation}). 
The task construction methods are divided into two categories: detection tasks and tracking tasks.
Detection tasks focus on assessing the ability of MLLMs to locate specific underwater objects, including searching for large targets such as sunken ships or aircraft wreckage, and smaller targets like scientific research robots. 
Tracking tasks focus on evaluating the ability of MLLMs to perform inspection and monitoring tasks underwater, including scenarios like pipeline inspection and platform approaches. 
To further investigate the performance in challenging environments, four representative tasks are conducted under low light deep-sea conditions. 
For each task, the starting position is selected to vary the difficulty, as tasks become more challenging when the starting point is far from the target, when the target is initially out of view, or when the agent initially faces away from the goal.
In our experiments, we follow a consistent initial positioning strategy: the first two starting positions are the same across all tasks to ensure reproducibility, while the third starting position is randomly set within the operational area to test how well the agent adapts to different starting conditions.

\begin{table*}[t]
\centering
\small
\scalebox{0.99}{
\begin{tabular}{@{}l*{10}{c}@{}}
\toprule
\multirow{3}{*}{Model} & \multicolumn{5}{c}{\makecell{Shallow Water Environment\\(High Illumination)}} & \multicolumn{5}{c}{\makecell{Deep Water Environment\\(Low Illumination)}} \\
\cmidrule(lr){2-6} \cmidrule(lr){7-11}
 & \multicolumn{2}{c}{\makecell{Multi-View\\Perception}} & \multicolumn{2}{c}{\makecell{Context-based\\Perception}} & \multirow{2}{*}{Avg} & \multicolumn{2}{c}{\makecell{Multi-View\\Perception}} & \multicolumn{2}{c}{\makecell{Context-based\\Perception}} & \multirow{2}{*}{Avg} \\
\cmidrule(lr){2-3} \cmidrule(lr){4-5} \cmidrule(lr){7-8} \cmidrule(lr){9-10}
 & Vision & +Sonar & Vision & +Sonar & & Vision & +Sonar & Vision & +Sonar & \\
\midrule
GLM-4.5V & 52.73 & \textbf{56.36} & 46.67 & 63.33 & \cellcolor{gray!10}54.77 & \underline{36.36} & 30.91 & 20.00 & \underline{33.33} & \cellcolor{gray!10}\underline{30.15} \\
GPT-4o-mini & 34.55 & 34.55 & 20.00 & 33.33 & \cellcolor{gray!10}30.61 & 14.55 & 20.00 & 3.33 & 6.67 & \cellcolor{gray!10}11.14 \\
Gemini-2.5-Flash & 29.09 & 30.91 & 50.00 & 33.33 & \cellcolor{gray!10}35.83 & 9.09 & 5.45 & 20.00 & 30.00 & \cellcolor{gray!10}16.14 \\
Qwen2.5-VL-7B & \textbf{58.18} & 43.64 & \underline{56.67} & \underline{70.00} & \cellcolor{gray!10}\underline{57.12} & 27.27 & 20.00 & 33.33 & \underline{33.33} & \cellcolor{gray!10}28.48 \\
Minicpm-4.5 & 52.73 & 43.64 & 36.67 & 23.33 & \cellcolor{gray!10}39.09 & 29.09 & 23.64 & \textbf{43.33} & 13.33 & \cellcolor{gray!10}27.35 \\
GPT-5.2 & \underline{56.36} & \textbf{56.36} & 36.67 & \textbf{73.33} & \cellcolor{gray!10}55.68 & \underline{36.36} & \underline{36.36} & 13.33 & \underline{33.33} & \cellcolor{gray!10}21.60 \\
Gemini-3-Pro & 45.45 & \underline{54.55} & \textbf{66.67} & \underline{70.00} & \cellcolor{gray!10}\textbf{59.17} & \textbf{40.00} & \textbf{38.18} & \underline{36.67} & \textbf{36.67} & \cellcolor{gray!10}\textbf{37.88} \\
\midrule
Human & 100.00 & 100.00 & 100.00 & 100.00 & \cellcolor{gray!10}100.00 & 94.55 & 98.18 & 86.67 & 90.00 & \cellcolor{gray!10}92.35 \\
\bottomrule
\end{tabular}}
\caption{Performance of perception tasks across different models and conditions. Values represent accuracy percentages (\%). Bold indicates the highest value in each column, underline indicates the second highest. Adding sonar means using both RGB and sonar images.}
\label{tab:main_results}
\end{table*}

\subsection{Evaluation Metrics}
\label{metrics}
\paragraph{Perception Task Evaluation.}
We evaluate model performance using exact match accuracy. Let $y_i$ denote the ground-truth answer and $\hat{y}_i$ represent the model’s predicted answer for the $i$-th sample.
\begin{align}
\text{Acc} = \frac{1}{N} \sum_{i=1}^N \mathbb{I}[\hat{y}_i = y_i]
\end{align}
For multiple-choice items, $y_i$ and $\hat{y}_i$ are treated as sets and equality requires an exact set match.

\paragraph{Decision Task Evaluation.}
% We evaluate decision tasks with a distance-based scoring scheme.
% Each episode ends when the agent issues \textsc{Stop} or the time limit $t_{\text{max}}$ is reached. For a task with $n$ evaluation points, let $\mathbf{p}_i$ denote the $i$-th ground-truth point. If the target is detected, we use the trajectory position closest to $\mathbf{p}_i$; otherwise, we use the final position of the agent. The Euclidean distance is $d_i=\lVert \hat{\mathbf{p}}_i-\mathbf{p}_i\rVert_2$, and the per-point score is:
We evaluate decision tasks using a distance-based scoring method. Each episode ends when the agent issues a \textsc{Stop} command or reaches the time limit $t_{\text{max}}$. For a task with $n$ evaluation points, let $\mathbf{p}_i$ be the $i$-th target location. If the target is detected, we use the closest position from the agent's trajectory to $\mathbf{p}_i$; otherwise, we use the agent's final position. The Euclidean distance is computed as $d_i = \lVert \hat{\mathbf{p}}_i - \mathbf{p}_i \rVert_2$, and the score for each point is defined as:
\begin{equation}
S_i=
\begin{cases}
100, & d_i \le \tau_1,\\[3pt]
100\,\dfrac{\tau_2-d_i}{\tau_2-\tau_1}, & \tau_1 < d_i \le \tau_2,\\[8pt]
0, & d_i > \tau_2,
\end{cases}
\end{equation}
where the distance thresholds are set to $\tau_1 = 30$ meters and $\tau_2 = 100$ meters by default.
The total score is a weighted sum  as $S_{\text{total}}=\sum_{i=1}^n w_i S_i$, where $w_i$ are task-specific weights\footnote{For a single-point task $w_1=1.0$; for two points $(w_1,w_2)=(0.6,0.4)$; for three points $(w_1,w_2,w_3)=(0.6,0.2,0.2)$.}.

\section{Experiments}

\subsection{Experimental Settings}
\label{sec:exp_setting}
To test MLLMs in underwater environments, we conduct experiments using various models. 
Among the open-source models, we assess MiniCPM-V-4.5 \citep{minicpm}, GLM-4.5V \citep{GLM-4.5V} and Qwen2.5-VL-7B \citep{qwenvl_2.5}. 
For proprietary models, we test GPT-4o-mini and GPT-5.2 \citep{gpt4o}, as well as Gemini-2.5-Flash and Gemini-3-Pro \citep{gemini}.
For perception tasks, we run each task three times. For decision-making tasks, each task involves three different starting points, with three runs per starting point. Laboratory human volunteers provide perception answers and operate robots via keyboards for decision tasks, where multiple trained participants each perform only one task to prevent bias from prior environment exposure.

\begin{table*}[t]
\centering
\small
\scalebox{0.99}{
\begin{tabular}{@{}l*{4}{c}c@{}}
\toprule
\multirow{2}{*}{Task} & \multicolumn{4}{c}{Model} & \multirow{2}{*}{Human} \\
\cmidrule(lr){2-5}
& GLM-4.5V & GPT-4o-mini & Gemini-2.5 & Qwen2.5-VL-7B \\
\midrule
\textbf{\makecell{Shallow Water Environment (High Illumination)}} & & & & \\
\midrule
Locate the robot    & $6.6_{\color{gray}\pm 19.83}$ & $\mathbf{8.9}_{\color{gray}\pm 10.1}$  & $0.0_{\color{gray}\pm 0.00}$ & $\underline{7.8}_{\color{gray}\pm 13.5}$ & 100 \\
Locate the oil drums        & $\underline{10.7}_{\color{gray}\pm 16.52}$ & $\mathbf{11.1}_{\color{gray}\pm 19.2}$ & $3.5_{\color{gray}\pm 6.0}$ & $5.7_{\color{gray}\pm 9.8}$ & 100 \\

Locate the electrical box   & $7.9_{\color{gray}\pm 17.08}$ & $\mathbf{36.6}_{\color{gray}\pm 21.9}$ & $\underline{15.9}_{\color{gray}\pm 27.4}$ & $8.7_{\color{gray}\pm 15.0}$ & 100 \\
Search for a sunken ship       & $5.9_{\color{gray}\pm 5.04}$ & $\underline{13.4}_{\color{gray}\pm 19.3}$ & $\mathbf{20.5}_{\color{gray}\pm 14.3}$ & $10.3_{\color{gray}\pm 10.3}$ & 100 \\

Search for the aircraft   & $\mathbf{25.0}_{\color{gray}\pm 6.10}$ & $\underline{16.9}_{\color{gray}\pm 17.8}$ & $11.7_{\color{gray}\pm 15.6}$ & $7.8_{\color{gray}\pm 10.0}$  & 100 \\

Inspect oil pipe         & $\mathbf{37.8}_{\color{gray}\pm 17.88}$ & $27.1_{\color{gray}\pm 23.6}$ & $18.3_{\color{gray}\pm 15.8}$ & $\underline{30.8}_{\color{gray}\pm 25.2}$   & 100 \\

Inspect the wind turbine      & $\underline{20.3}_{\color{gray}\pm 28.89}$ & $13.9_{\color{gray}\pm 14.33}$  & $\mathbf{25.1}_{\color{gray}\pm 22.1}$ & $14.7_{\color{gray}\pm 17.0}$  & 100 \\

Docking     & $14.9_{\color{gray}\pm 13.20}$ & $\underline{19.2}_{\color{gray}\pm 33.28}$ & $\mathbf{19.4}_{\color{gray}\pm 33.6}$ & $8.3_{\color{gray}\pm 7.2}$  & 100 \\
\rowcolor{gray!10} \textbf{Average} & $\underline{16.1}_{\color{gray}\pm 15.6}$ & $\mathbf{18.4}_{\color{gray}\pm 19.9}$ & $14.4_{\color{gray}\pm 16.1}$ & $11.8_{\color{gray}\pm 13.7}$  & 100 \\
\midrule
\textbf{\makecell{Deep Water Environment (Low Illumination)}} & & & & \\
\midrule
Locate oil drums      & $\mathbf{10.6}_{\color{gray}\pm 21.35}$ & $\underline{5.6}_{\color{gray}\pm 9.69}$  & $0.0_{\color{gray}\pm 0.0}$ & $0.0_{\color{gray}\pm 0.0}$  & 40.8  \\
Search for a sunken ship & $2.9_{\color{gray}\pm 2.16}$ & $\mathbf{12.8}_{\color{gray}\pm 14.48}$ & $\underline{8.2}_{\color{gray}\pm 14.1}$ & $3.4_{\color{gray}\pm 5.8}$  & 100 \\
Inspect the oil pipe   & $\mathbf{32.5}_{\color{gray}\pm 5.86}$ & $15.8_{\color{gray}\pm 15.5}$ & $6.6_{\color{gray}\pm 11.4}$ & $\underline{21.7}_{\color{gray}\pm 25.3}$ & 78.2 \\
Inspect the wind turbine      & $0.0_{\color{gray}\pm 0.0}$ & $\mathbf{25.1}_{\color{gray}\pm 16.0}$ & $\underline{10.6}_{\color{gray}\pm 10.0}$ & $0.4_{\color{gray}\pm 0.6}$  & 100 \\
\rowcolor{gray!10} \textbf{Average} & $\underline{11.5}_{\color{gray}\pm 7.3}$ & $\mathbf{14.8}_{\color{gray}\pm 13.9}$ & $6.4_{\color{gray}\pm 8.8}$ & $6.4_{\color{gray}\pm 8.4}$ & 69.6 \\
\bottomrule
\end{tabular}
}
\caption{Performance in decision tasks requiring autonomous completion by MLLM-driven agents.}
\label{tab:task_performance}
\end{table*}

\begin{figure*}[t]
\centering
\begin{minipage}[t]{0.49\textwidth}
    \centering
    \includegraphics[width=0.99\linewidth]{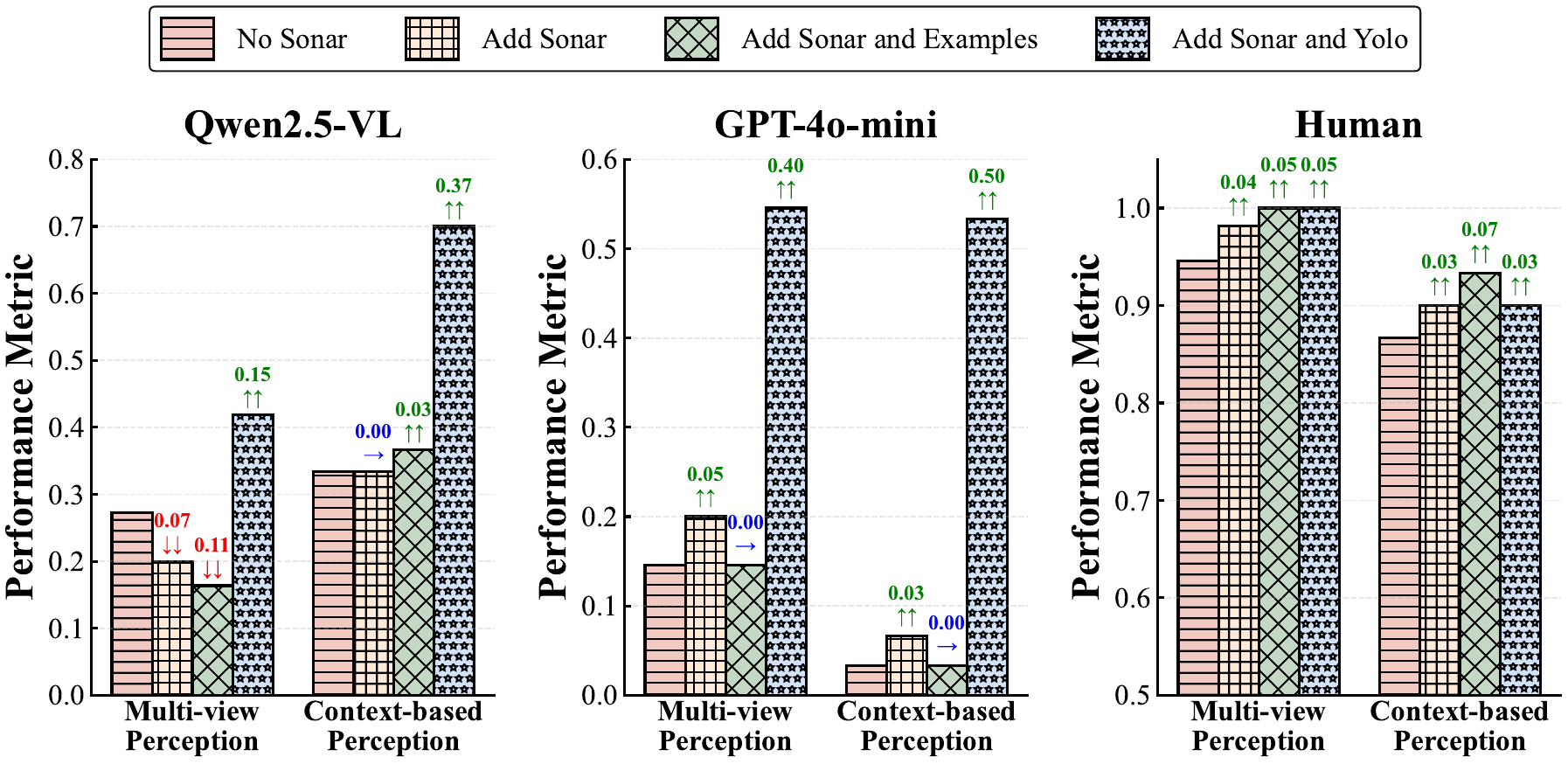}
    \par\vspace{0.2cm}
    \caption{Performance comparison between human and MLLMs after adding sonar and sonar reference examples for objects in deep water environments.}
    \label{fig:sonar}
\end{minipage}
\hspace{0.02\textwidth}
\begin{minipage}[t]{0.45\textwidth}
    \centering
    \includegraphics[width=0.72\linewidth]{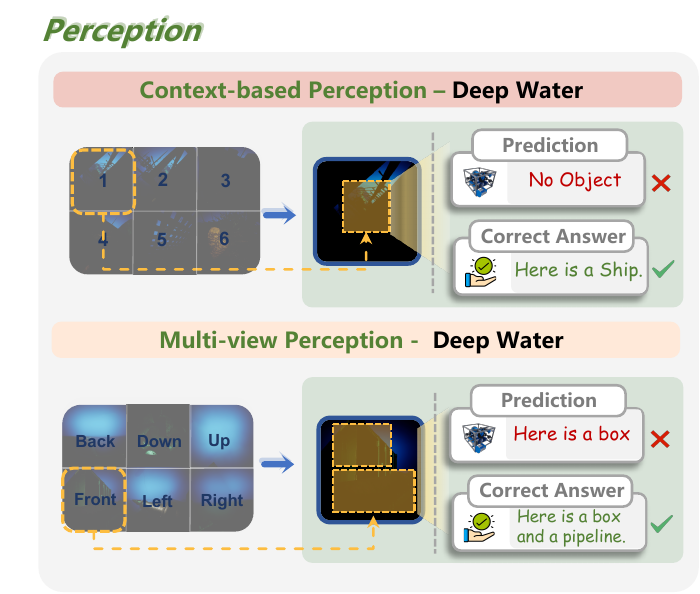}
    \par\vspace{0.2cm}
    \caption{Case analysis in perception tasks. Agents are susceptible to errors under challenging conditions such as under low-light, multi-object, or occluded conditions.}
    \label{fig:case1}
\end{minipage}
\end{figure*}

\subsection{Main Results}

\paragraph{Perception Results.}
The results for perception tasks are summarized in Table~\ref{tab:main_results}. In shallow, well-illuminated water environments, Gemini-3-Pro achieves the strongest overall performance among the evaluated MLLMs, with an average accuracy of 59.17\%, followed by Qwen2.5-VL-7B (57.12\%) and GLM-4.5V (54.77\%). Under deep water conditions with low illumination, all models exhibit significant performance degradation, though Gemini-3-Pro emerges as the most robust (37.88\% average accuracy), followed by GLM-4.5V (30.15\%), Qwen2.5-VL-7B (28.48\%), and MiniCPM-V-4.5 (27.35\%). Notably, incorporating sonar data does not consistently improve performance across models or tasks. The performance gap between MLLMs and human level indicates substantial room for improvement in underwater perception tasks and degraded visual inputs.

\paragraph{Decision Results.}
Performance on decision tasks is shown in Table~\ref{tab:task_performance}. Several tasks resulted in zero scores, indicating extreme difficulty due to small object size or time constraints. GPT-4o-mini achieves the best average performance in both shallow (18.4\%) and deep water (14.8\%) environments, with GLM-4.5V ranking second under shallow conditions (16.1\%) and deep water conditions (11.5\%). Performance declines markedly in deep water, where Gemini-2.5-Flash and Qwen2.5-VL-7B both average 6.4\%. 
Human performance substantially outperforms all models, reaching 100\% in shallow water and 69.6\% in deep water, underscoring the gap between MLLM-driven agents and humans.

\begin{figure*}[t]
\centering
\begin{minipage}{0.99\textwidth}
\vspace{-0.1in}
    \centering
    \includegraphics[width=\textwidth]{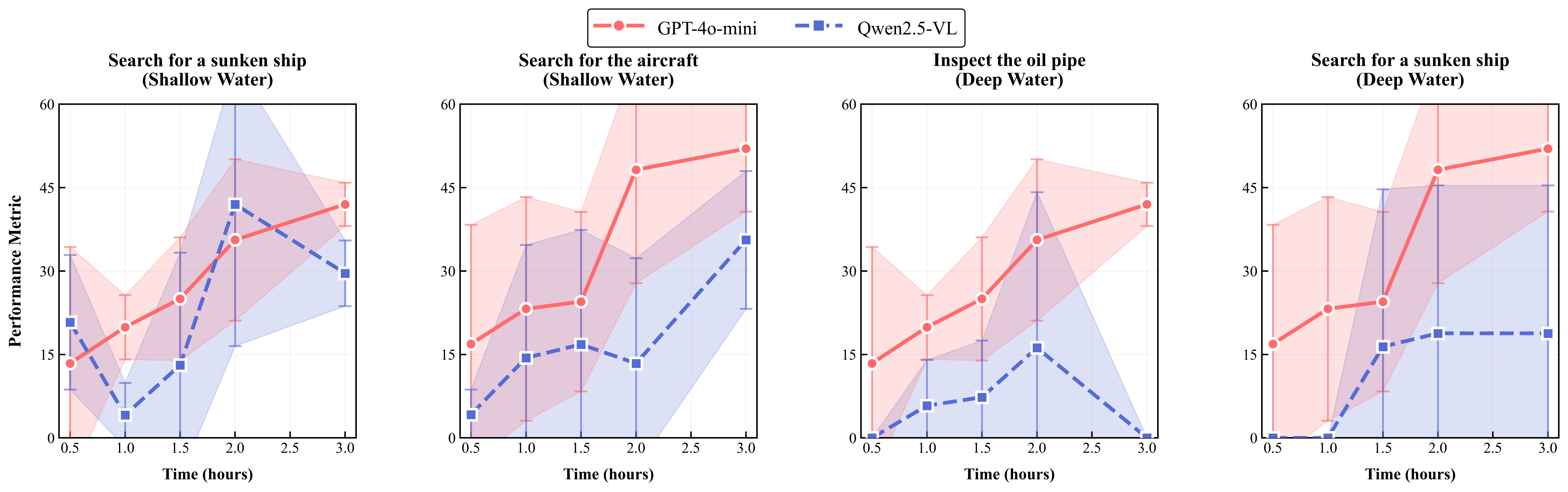}
    \captionof{figure}{Scaling analysis performance over time in decision tasks.}
    \label{fig:scaling}
\end{minipage}

\vspace{0.1cm}

\begin{minipage}{0.99\textwidth}
    \centering
    \includegraphics[width=\textwidth]{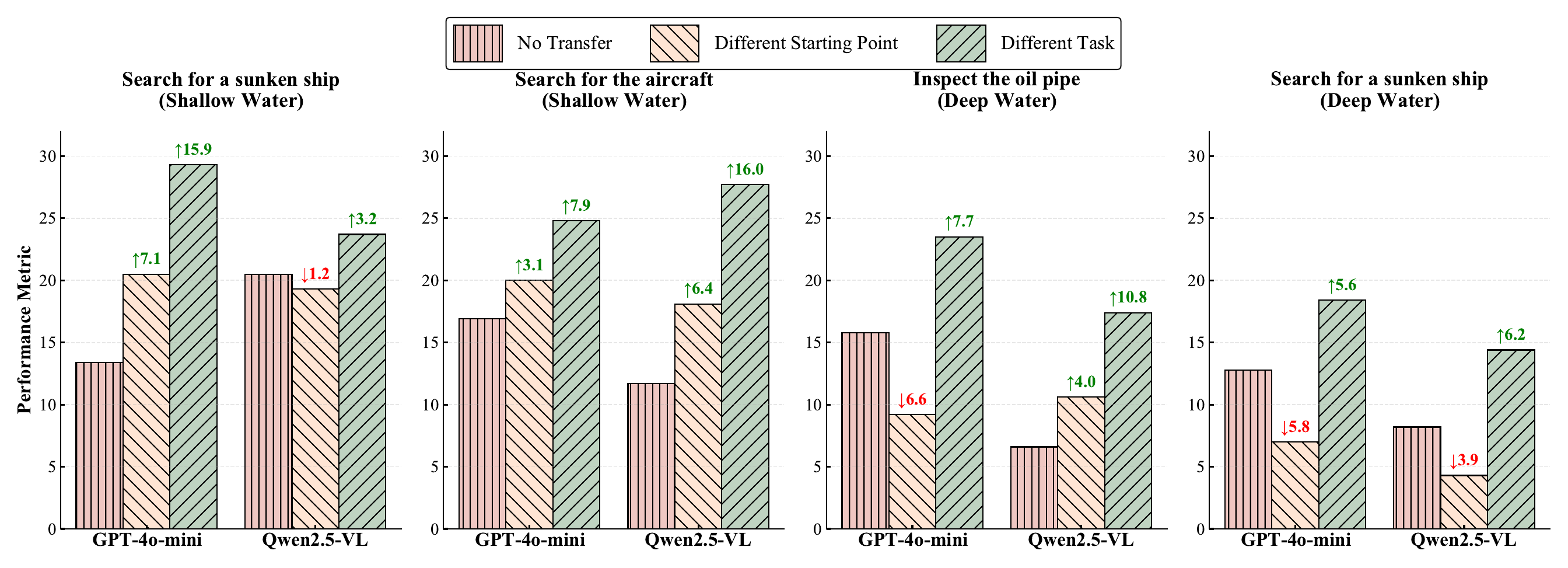}
    \captionof{figure}{Impact of different memory transfer mechanisms on model performance.}
    \label{fig:transfer}
\end{minipage}
\label{fig:scaling_transfer}
\vspace{-0.2cm}
\end{figure*}

\begin{figure*}[t]
\centering
\includegraphics[width=0.99\textwidth]{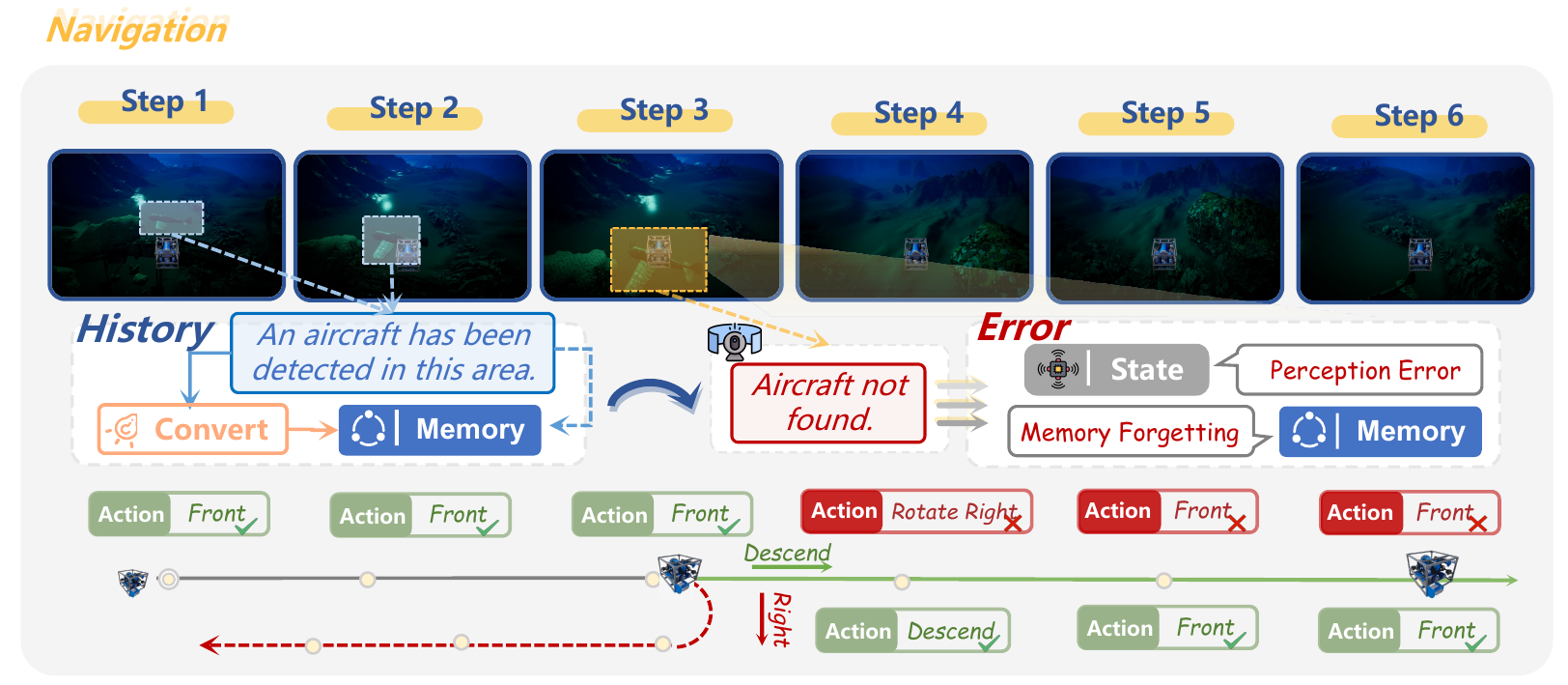}
\caption{
Case analysis in decision tasks.
}
\label{fig:case2}
\end{figure*}

\subsection{Analysis}
\label{analysis}

\paragraph{MLLM agents struggle to exploit sonar data for enhanced underwater perception, in stark contrast to humans who leverage it effectively.}
To investigate the role of sonar data in deep-water environments, we compare the performance of human experts with the two MLLMs on perception tasks.
Specifically, we either let the models directly comprehend sonar images or provide them with human-annotated interpretations as prompts.
As shown in Figure~\ref{fig:sonar}, human experts consistently benefit from incorporating sonar data across tasks. 
By contrast, MLLMs exhibit only limited gains when using raw sonar images, and this gap becomes even more pronounced when reference sonar images of each object are introduced. As shown in Table~\ref{tab:error_analysis}, \textbf{sonar improves presence detection for both models and humans, but does not help models with object identification.} This limitation likely stems from current MLLMs' fundamental difficulty in interpreting sonar imagery and underwater perceptual data \citep{sonardata, zheng2023marinegpt, NAUTILUS, aubard2025sonar}. 
Notably, when employing a YOLO model \citep{yolo} specifically trained on sonar data as auxiliary perception tools, we observe performance improvements, suggesting that specialized vision models may currently outperform general-purpose MLLMs in sonar data usage.

\paragraph{Extended exploration boosts agent knowledge and task performance, following a scaling law that eventually plateaus.}
We analyze the relationship between navigation performance and operational duration using the representative MLLMs, across both shallow- and deep-water scenarios. The performance was evaluated over durations of 0.5, 1, 1.5, 2, and 3 hours.
As shown in Figure~\ref{fig:scaling}, performance initially improves with longer operation time, consistent with prior studies on test-time scaling \citep{survey_tts1, survey_tts2}, but eventually plateaus. 
This plateau reflects inherent limitations in perception, memory, and reasoning, as well as a lack of intrinsic curiosity to explore new regions.
These findings underscore the need to improve both fundamental MLLM capabilities and agent strategies, such as enhanced memory and long-horizon planning, to break through performance ceilings in embodied environments.

\paragraph{Memory transfer enables agents to leverage past experience to tackle new challenges.}
We investigate whether knowledge and experience accumulated from previous tasks \citep{exp5,exp6,exp1,exp2} can enhance performance in new tasks. Specifically, we explore using agents' previously explored trajectories as experiential input. Experiments are conducted in both shallow water and deep water environments, evaluating two transfer conditions: within-task transfer (different starting points) and cross-task transfer (different but related tasks).
As shown in Figure~\ref{fig:transfer}, memory transfer improves decision-making performance in shallow water environments under both transfer conditions. However, in the more challenging deep water environment, only cross-task transfer demonstrates stable performance improvements, while within-task transfer shows limited benefits. This suggests that more appropriate prior experiences provide more robust guidance under perceptually degraded conditions.
Transfer learning helps compensate for perceptual limitations by providing informed priors about environmental structure and effective navigation strategies. These findings underscore the importance of developing adaptive memory retrieval mechanisms that can selectively leverage relevant past experiences to enhance decision-making in autonomous underwater agents operating under diverse environmental conditions.

\paragraph{Case analysis.}
We present perception case analyses and illustrate failure cases in Figure~\ref{fig:case1}, where failures mainly arise from: \textbf{(1) Occlusions}, where targets are partially blocked; \textbf{(2) Multi-object Scenes}, causing identification and localization ambiguities; and \textbf{(3) Low Illumination}, which severely reduces vision-based perception accuracy.
Figure~\ref{fig:case2} shows common decision task failures, primarily arising from: \textbf{(1) Perception Errors}, where inaccurate detection leads to wrong actions; and \textbf{(2) Memory Forgetting}, where the agent cannot retain crucial past information, such as visited locations or previous decisions. For a more detailed error analysis, see Appendix~\ref{app:error}.

\paragraph{Sim-to-Real transfer analysis.}
To bridge simulation and reality, we deploy physical reference objects from OceanGym's object modeling into a real marine environment, enabling correlation between simulated and real-world performance. An AUV equipped with a sonar data acquisition system collects sonar measurements. As shown in Figure~\ref{fig:real-world_case}, \textbf{the YOLO model trained in simulation enhances GPT-4o-mini's ability to interpret real-world sonar data}. However, it exhibits limited generalization to objects not included in the simulation.
As illustrated in Figure~\ref{fig:sim2real},\textbf{ aligning the model's output actions with real operational commands allows prompts from the simulated environment to directly control real AUVs}. Furthermore, with environment memory that mirrors reality, the model can rapidly follow language commands to locate or reach designated targets.

% \paragraph{Discusses and Limitations of {\bench}.}
% \label{discuss}
% {\bench} offers a versatile testbed for underwater embodied agents, though it cannot fully replicate real-world conditions as factors like currents, salinity, marine life, and sonar noise remain imperfectly modeled. 
% Despite these constraints, {\bench} supports synthetic data generation and facilitates reinforcement learning with rich feedback, and serves as a sim-to-real bridge for deploying models on AUVs  (See Appendix \ref{app:discuss}).

%strained models on real AUVs.  More detailed are in  Appendix \ref{app:discuss}.

\section{Related Work}
Embodied intelligence relies on simulation platforms across terrestrial~\citep{Matterport3D,House3D,habitat}, aerial~\citep{airsim,CityNav,OpenUAV}, and marine domains~\citep{holoocean,oceansim,MarineGym}. The emergence of MLLMs has advanced embodied agents by integrating visual understanding~\citep{gpt4o,qwenvl_2.5,gemini}. Existing embodied benchmarks cover indoor~\citep{r2r,House3D}, urban~\citep{Touchdown, EmbodiedCity}, aerial~\citep{OpenFly, FlightGPT}, and real-world~\citep{UrbanVideo, koh2024visualwebarena} settings, yet a unified benchmark for underwater multimodal embodied agents remains absent. More related work is provided in Appendix~\ref{app:works}.

\section{Conclusion}
We introduce {\bench}, the first benchmark for underwater embodied agents, with the goal of bridging the gap between simulation and reality. {\bench} presents marine challenges through a high-quality platform that features realistic environments, diverse sensing modalities, and a wide range of underwater tasks. Experiments reveal limitations of current MLLMs in underwater settings.

% \clearpage

\section*{Limitations}
\label{limitations}
While {\bench} provides a valuable testbed for underwater embodied agents, several limitations should be acknowledged.
OceanGym leverages Unreal Engine (UE) 5.3 \citep{unrealengine} for realistic underwater environment rendering and physical simulation, while utilizing HoloOcean's \citep{holoocean} cluster-based multipath ray-tracing algorithm to simulate multibeam sonar.
Although UE plugins, as exemplified by those shown in Figure~\ref{fig:vid_frames}, can be used to simulate water flow, buoyancy, lighting, water interaction etc, they cannot fully replicate the real underwater environment, as factors such as ocean currents, salinity, marine life, and geological changes are not accurately captured.
Future work may leverage generative models \citep{ball2025genie} or physics-informed machine learning to incorporate these complexities.
The optical and sonar images still differ from those in the real world, particularly since sonar simulation introduces errors. 
In addition, the environment is large and requires considerable computational resources.
Moreover, the memory methods employed in our framework are rudimentary, lacking more advanced designs.
These limitations highlight opportunities for future work to expand task coverage, improve physical realism, and optimize computational efficiency, and enhance memory mechanisms.
% \subsection*{Ethics Statement}
% This research is conducted in strict compliance with established ethical guidelines and best practices in scientific research. 
% All data employed in this study are obtained from publicly accessible datasets, with no utilization of proprietary or confidential information. Proper and accurate citations are provided for all data sources referenced throughout this paper.
% We emphatically advise all users to maintain the highest ethical standards when utilizing our dataset, ensuring principles of fairness, transparency, and responsibility in their research applications. 
% Any use of the dataset that may potentially cause harm or adversely affect societal welfare is expressly prohibited.

\section*{Acknowledgements}
We would like to express our sincere gratitude to the anonymous reviewers for their thoughtful and constructive feedback. This work was supported by the New Generation Artificial Intelligence-National Science and Technology Major Project (2025ZD0122802), National Natural Science Foundation of China (No.62576307), the Fundamental Research Funds for the Central Universities (226-2023-00138), CCF-Tencent Open Research Fund, the Yongjiang Talent Introduction Programme (2021A-156-G), and the Information Technology Center and State Key Lab of CAD\&CG, Zhejiang University.

% \subsection*{Reproducibility Statement}
% We provide data from our benchmark under file size limitation, along with the corresponding evaluation code, in the supplementary materials. Detailed descriptions of the environment setup and data construction procedures are available in \S~\ref{sec:env}, \S~\ref{sec:task1} and \S~\ref{sec:task2}. Additional data details and comprehensive benchmark statistics can be found in Appendix~\ref{app:perception} and Appendix~\ref{app:navigation}. 
% Specific configurations of the tested models are documented in Section~\ref{sec:exp_setting}.

\bibliography{custom}

\appendix
% \clearpage

\section*{Appendix}
\section{The Use of Large Language Models (LLMs)}
We confirm that LLMs are used only as an auxiliary tool to assist in refining wording and sentence structure. 
Their application in experiments is strictly confined to scientific research purposes, and all such uses have been clearly documented in the Experimental Settings. No additional reliance on LLMs has been involved in this work.

\begin{figure*}[h]
\centering
\includegraphics[width=0.99\textwidth]{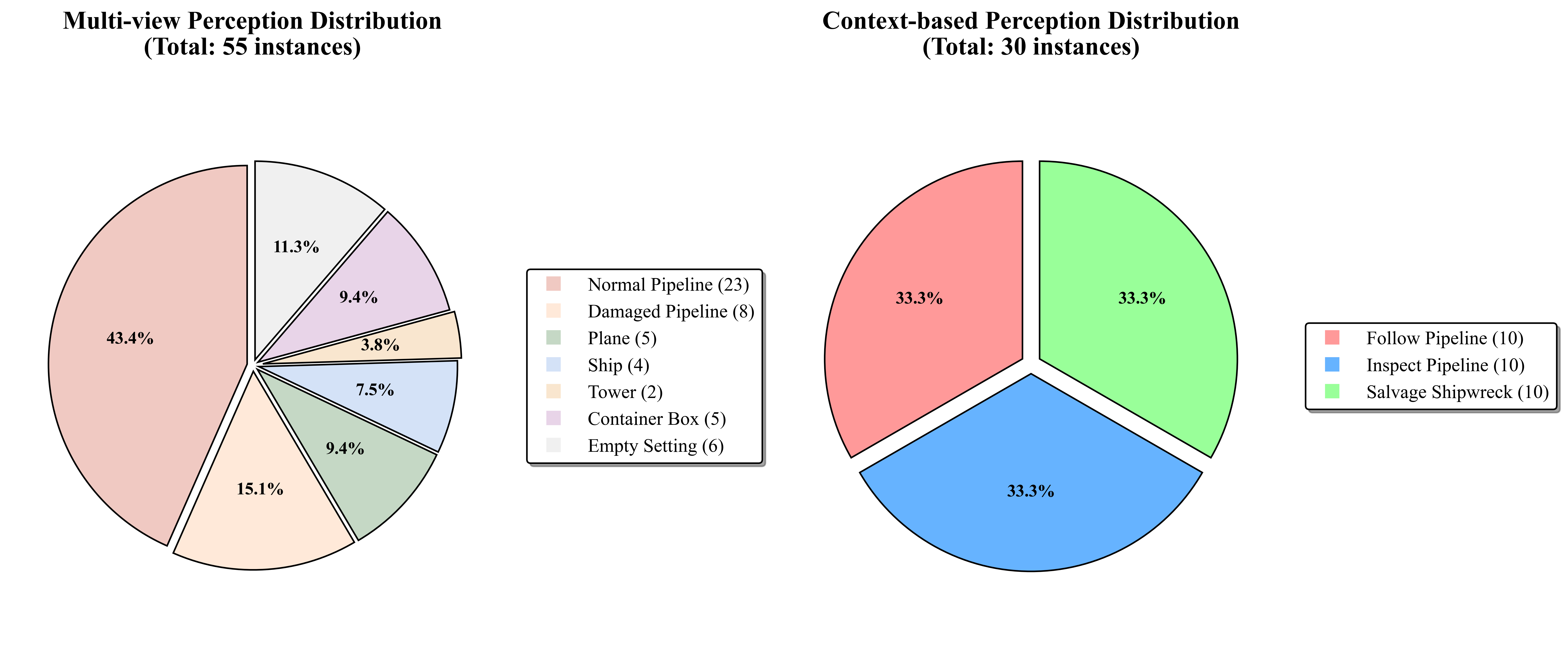}
\caption{Statistics of perception tasks.}
\label{fig:pie}
\end{figure*}

\section{More Related Works}
\label{app:works}
\paragraph{Embodied Simulations.} Embodied intelligence describes artificial intelligence systems whose intelligent behavior emerges through continuous physical and sensory interactions with the environment \citep{embodiedai1, embodiedai2, embodiedai3}. Simulation platforms are essential for advancing such systems across ground, aerial, and marine domains \citep{aligningcyberspacephysical, survey_for_robot, underwaterroboticsimulatorsreview}. In ground applications, platforms like Matterport3D \citep{Matterport3D}, House3D \citep{House3D}, and Habitat \citep{habitat} provide realistic indoor and outdoor environments for navigation, scene understanding, and human-robot interaction research. Aerial robotics benefits from simulators such as AirSim \citep{airsim}, CityNav \citep{CityNav}, and OpenUAV \citep{OpenUAV}, which offer high-fidelity simulations with accurate physics and sensor models. Similarly, in the marine domain, simulation platforms such as HoloOcean \citep{holoocean}, OceanSim \citep{oceansim}, and MarineGym \citep{MarineGym} provide specialized capabilities for modeling hydrodynamic effects and underwater dynamics. With the development of embodied intelligence, an increasing variety of simulation environments \citep{kolve2017ai2, puig2018virtualhome, xiang2020sapien, ThreeDWorld, li2021igibson, nasiriany2024robocasa, HAZARD, EmbodiedWeb} have emerged to support specific tasks, needs, or scenarios.

\paragraph{MLLM-driven Embodied Agents.} Building upon the rapid advancement of LLMs \citep{gpt4,llama2023,vicuna2023,qwen}, the emergence of MLLMs \citep{gpt4o,qwenvl_2.5,llama3.2, liu2024llavanext, gemini, GLM-4.5V, internvl35} has further strengthened agent capabilities by incorporating visual understanding for multimodal perception. Despite impressive results in various agent applications \citep{agent-survey1, agent-survey2}, MLLM-driven agents still face substantial challenges in real-world and simulated embodied environments. 
Key difficulties persist in spatial cognition \citep{rephrase,embspatial,eyes_wide,empirical_analysis,Reefknot,RynnEC,MMSI-Bench,STI-Bench}, task planning \citep{chen2023robogpt, huang2023embodied,
zhou2024navgpt}
, object navigation \citep{visionlanguagenavigation,goatbench,BEDI,NavBench,EmbodiedEval}, and robotic manipulation \citep{vlmbench,EmbodiedBench,chainofmodalitylearning}. 
To evaluate agent capabilities, embodied benchmarks have been developed across diverse settings, including indoor \citep{r2r,House3D}, urban \citep{Touchdown, NuScenes, Talk2Nav, EmbodiedCity}, aerial \citep{AeroVerse, OpenFly, FlightGPT}, specialised \citep{zheng2022vlmbench, luo2023fmb, song2024towards, BEHAVIOR}, and real-world \citep{UrbanVideo, koh2024visualwebarena, MME} scenarios.

\section{Applications and Future Directions of {\bench}}
\label{app:app}

\paragraph{Competitive Arena for Foundation Models.}
{\bench} provides a rigorous testbed for evaluating foundation models and embodied agent frameworks. Future work can leverage {\bench} to optimize prompt design, enhance base model capabilities, improve agent design, and assess trained or fine-tuned models across diverse underwater scenarios.

\paragraph{Memory Mechanism Evaluation.}
{\bench} enables systematic evaluation of memory mechanisms in embodied agents. While prior works in aerial and ground domains have explored memory summarized as natural language descriptions, more advanced multimodal memory methods have since been developed. This motivates our choice of sliding window as a baseline. Future work can leverage {\bench} to improve memory utilization and compare different memory architectures for underwater tasks, such as retrieval-based memory or hierarchical memory.
Beyond memory, future work can also explore standardized baseline systems such as classical SLAM combined with planning modules.

\paragraph{Underwater Data Synthesis Platform.}
The framework enables the generation of synthetic simulation data to improve both perception and decision-making capabilities of autonomous agents. This data can be effectively used for training models, and, given the inherent difficulties in acquiring real-world underwater data, it serves as a valuable supplement to scarce real-world datasets.

\paragraph{Reinforcement Learning Testbed with Comprehensive Feedback Signals.}
{\bench} offers an environment for reinforcement learning research with comprehensive feedback signals. Researchers can leverage its diverse reward mechanisms, customizable configurations, and real-time metrics to benchmark RL algorithms under varying underwater conditions.

\begin{figure*}[ht]
\centering
\includegraphics[width=1\textwidth]{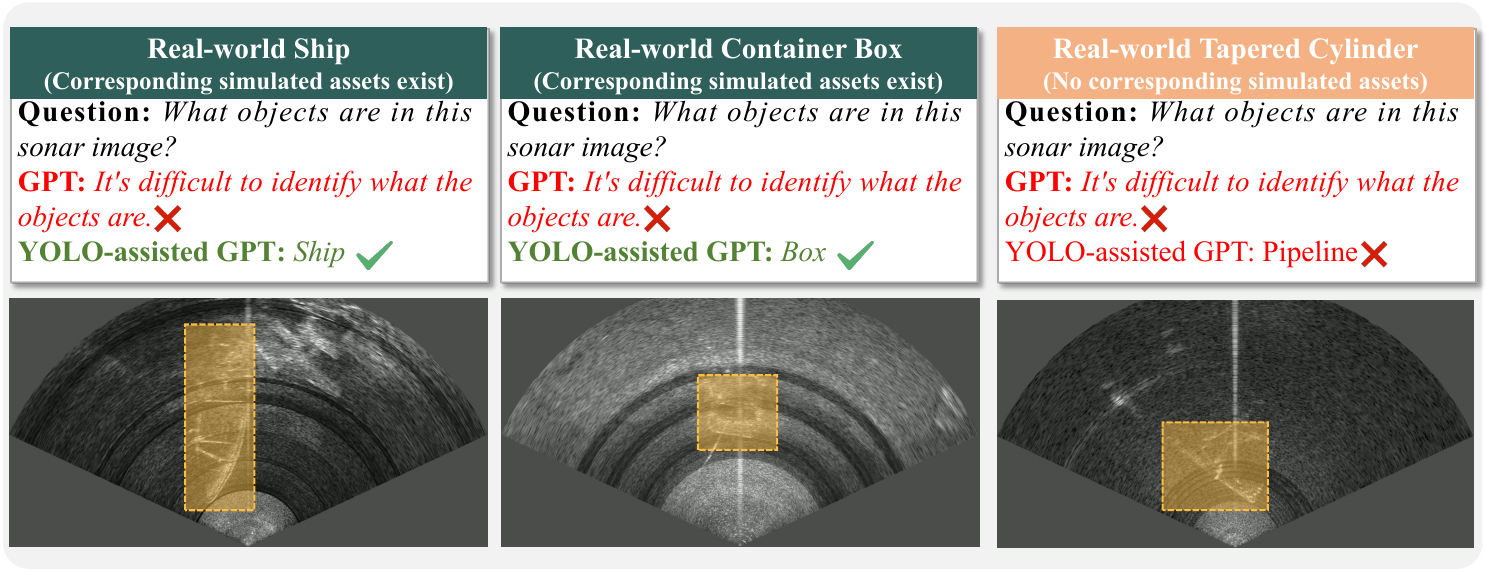}
\caption{
We evaluate whether YOLO models trained in simulated environments can enhance real-world performance by testing them on actual sonar data. The results demonstrate that while the YOLO-assisted GPT-4o-mini approach yields measurable performance improvements for certain objects modeled in {\bench}, the models exhibit limited generalization capability for objects not included in the simulation.
}
\label{fig:real-world_case}
\end{figure*}

\begin{figure*}[ht]
\centering
\includegraphics[width=1\textwidth]{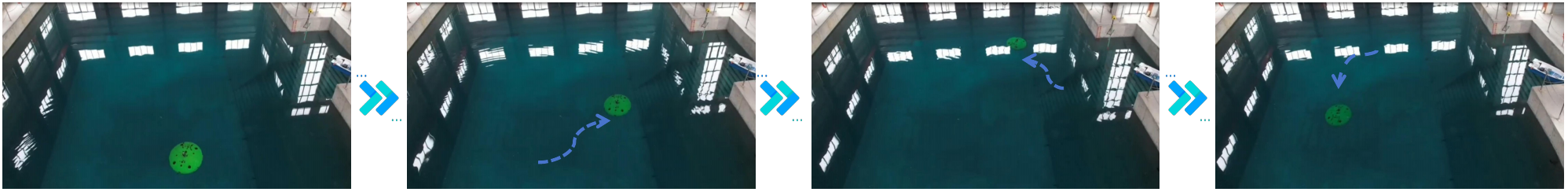}
\caption{
By matching the model's output actions with real operation commands, such as those standardized in the MOOS-IvP or ROS, the same prompts used in the simulated environment can directly control real AUVs. Moreover, using environment memory that mirrors reality as input enables the model to rapidly follow language instructions to reach or locate target points. In the picture, using simulated environment prompts and memory to help the real AUV locate three pre-set underwater objects as target points.
}
\label{fig:sim2real}
\end{figure*}

\paragraph{Sim-to-Real Bridge.}
The platform represents an initial step toward enabling the transfer of trained models to real-world AUVs, serving as a promising yet preliminary link between simulation and deployment. As illustrated in Figure~\ref{fig:real-world_case}, YOLO models trained in {\bench}'s simulated environment demonstrate measurable performance improvements when tested on real sonar data, though generalization remains limited for objects not included in the simulation. Furthermore, Figure~\ref{fig:sim2real} shows that by aligning model outputs with standardized robotic operating commands, the same prompts and environment memory used in simulation can directly guide real AUVs to follow language instructions and locate target points. Through this initial exploration of connecting virtual testing with real-world deployment, {\bench} offers a foundation for reducing reliance on expensive and hazardous field trials, accelerating development cycles, and enhancing the reliability and robustness of autonomous underwater systems.

\section{Perception Task Statistics}
\label{app:perception}
Figure~\ref{fig:pie} presents the statistical distribution of perception settings in our dataset. 
The dataset comprises 85 data sets in total: 55 for Multi-view Perception and 30 for Context-based Perception.
The 55 Multi-view Perception sets are distributed as follows: 23 normal pipeline scenes, 8 damaged pipeline scenes, 5 plane-related scenes, 4 ship scenes, 2 tower scenes, 5 container box scenes, and 6 scenes with no dominant object.
The 30 Context-based Perception sets are evenly divided into three sub-tasks, each with 10 sets. These sub-tasks involve the agent following pipelines, inspecting pipelines for potential damage, and scanning shipwreck areas.
% As shown in Table~\ref{tab:more_tasks}, we test the model's perception capabilities on additional scenes, and the overall scores remain consistently stable.

% \input{table/more_tasks}

\begin{figure*}[htbp]
\centering
\includegraphics[width=0.98\textwidth]{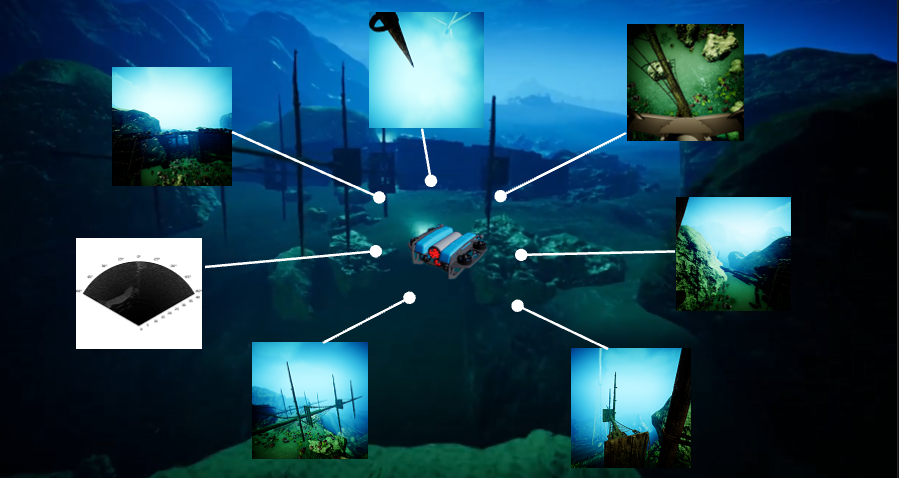}
\caption{A state case of a decision-making task.}
\label{fig:perspective}
\vspace{-0.5cm}
\end{figure*}

\begin{figure*}[ht]
    % \vspace{-0.5in}
    \centering
    \begin{minipage}[t]{0.24\textwidth}
        \centering
        \includegraphics[width=\textwidth]{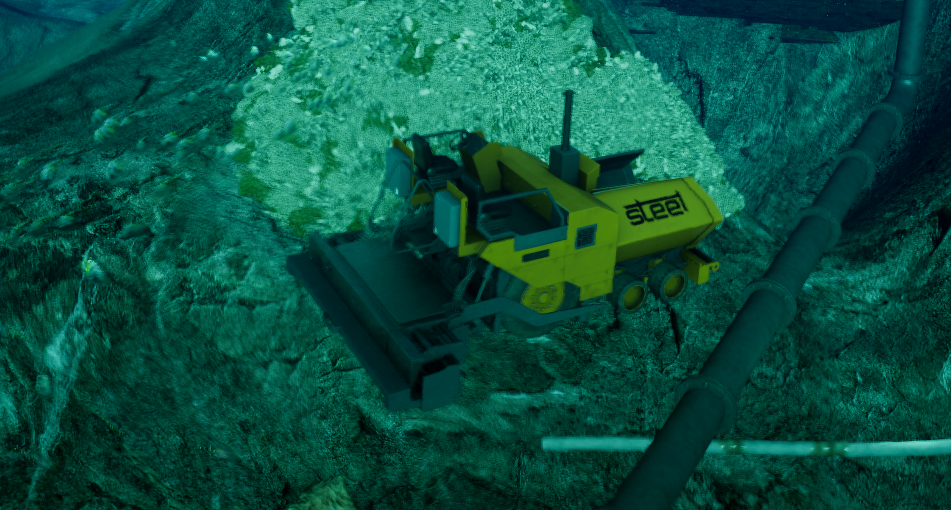}
        \caption{ Target object for the ``Locate the robot'' task.}
    \end{minipage}
    \hfill
    \begin{minipage}[t]{0.24\textwidth}
        \centering
        \includegraphics[width=\textwidth]{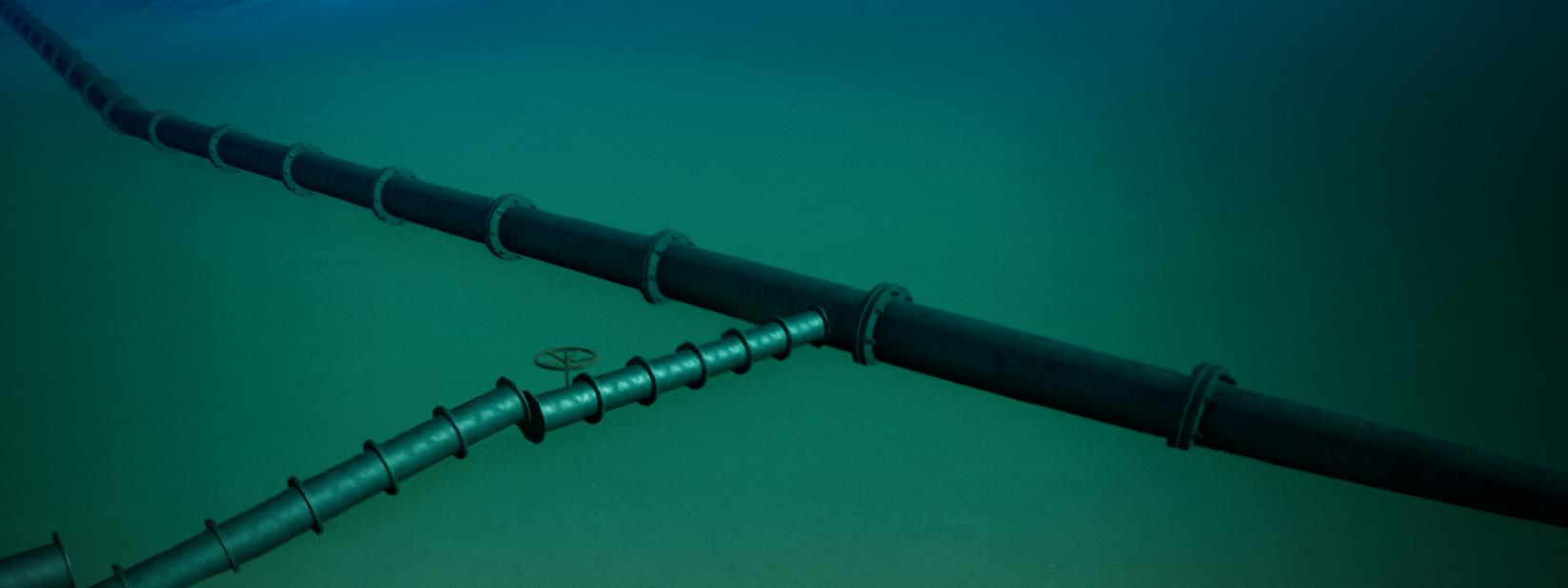}
        \caption{ Target object for the ``Inspect the oil pipe'' task.}
    \end{minipage}
    \hfill
    \begin{minipage}[t]{0.24\textwidth}
        \centering
        \includegraphics[width=\textwidth]{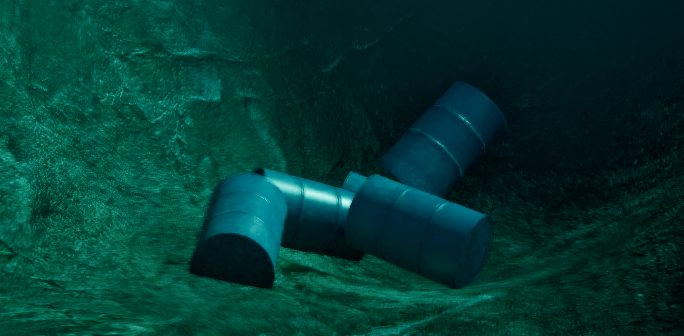}
        \caption{ Target object for the ``Locate oil drums'' task.}
    \end{minipage}
    \hfill
    \begin{minipage}[t]{0.24\textwidth}
        \centering
        \includegraphics[width=\textwidth]{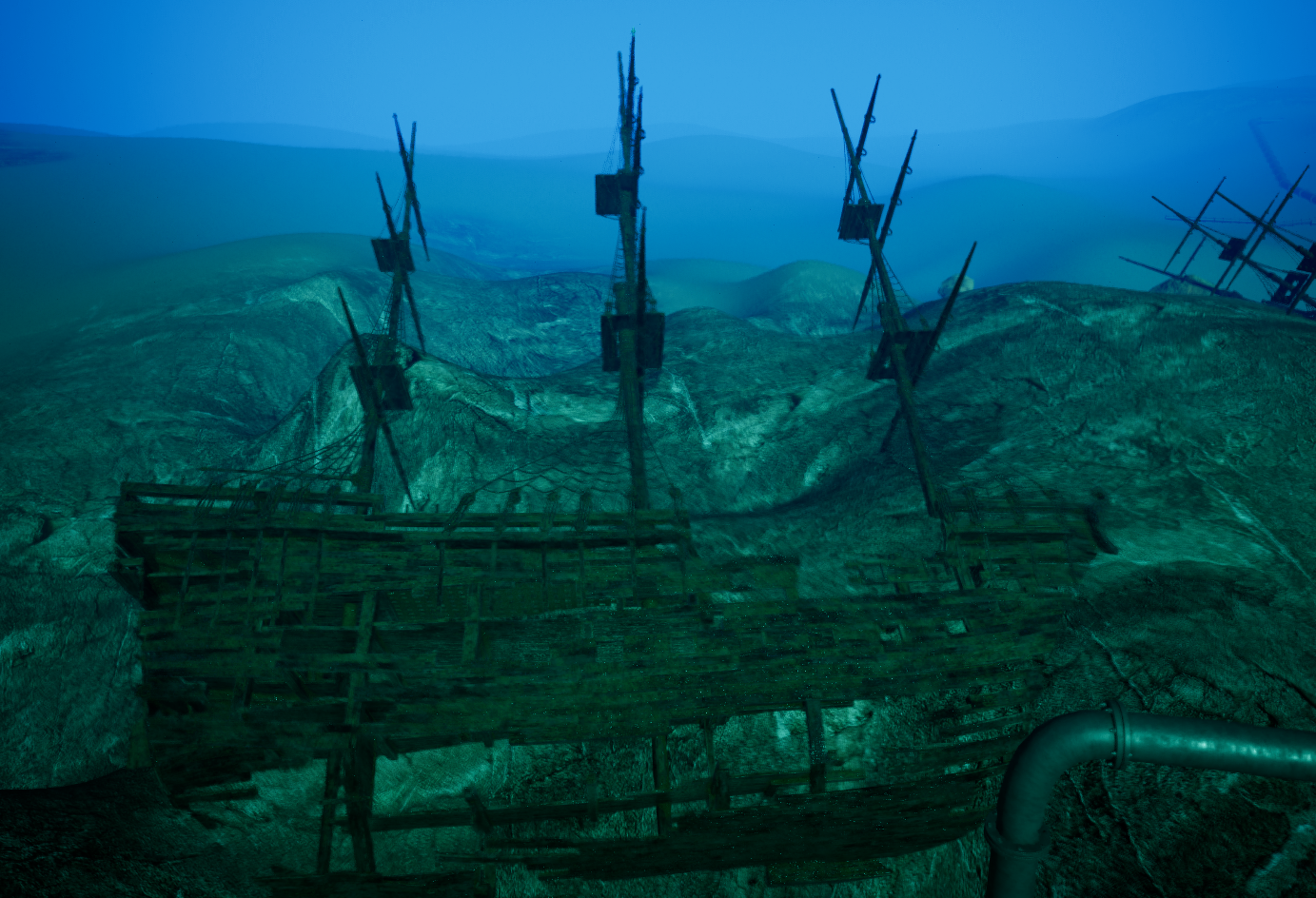}
        \caption{ Target object for the ``Search for a sunken ship'' task.}
    \end{minipage}
    \vspace{0.3cm}
    \begin{minipage}[t]{0.24\textwidth}
        \centering
        \includegraphics[width=\textwidth]{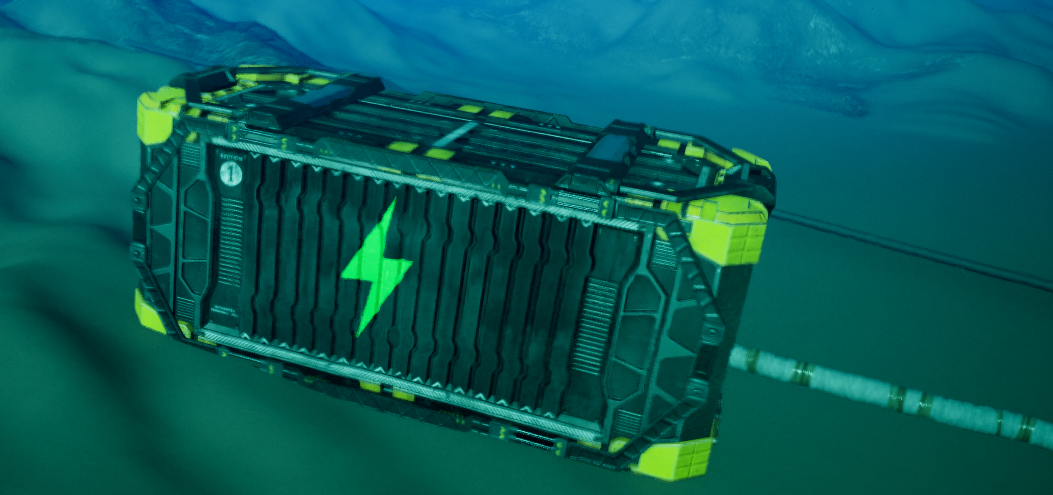}
        \caption{ Target object for the ``Locate the electrical box'' task.}
    \end{minipage}
    \hfill
    \begin{minipage}[t]{0.24\textwidth}
        \centering
        \includegraphics[width=0.2\textwidth]{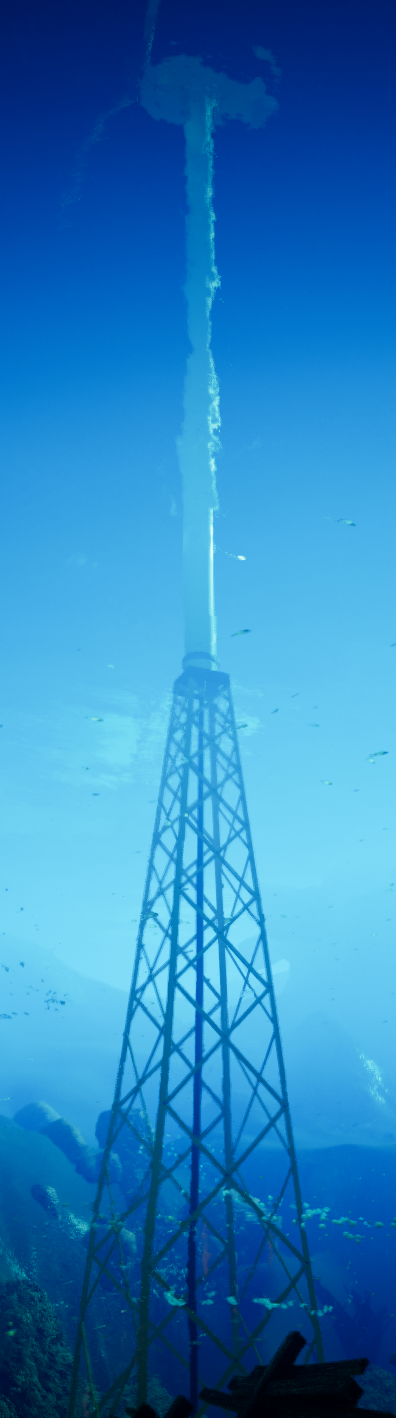} % Assuming this is shorter
        \caption{ Target object for the ``Inspect the wind turbine'' task.}
    \end{minipage}
    \hfill
    \begin{minipage}[t]{0.24\textwidth}
        \centering
        \includegraphics[width=\textwidth]{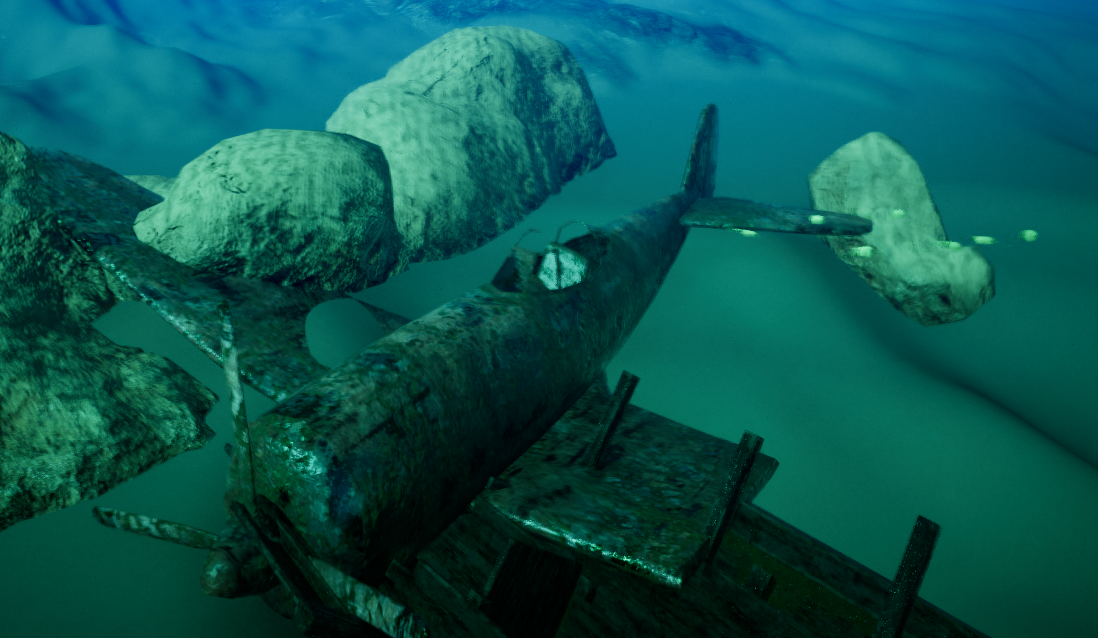}
        \caption{ Target object for the ``Search for the aircraft'' task.}
    \end{minipage}
    \hfill
    \begin{minipage}[t]{0.24\textwidth}
        \centering
        \includegraphics[width=\textwidth]{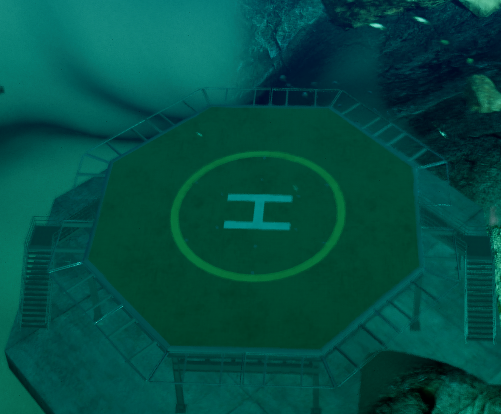}
        \caption{ Target object for the ``Docking'' task.}
    \end{minipage}
    \label{fig:tasks}
    % \vspace{-0.5cm}
\end{figure*}

\section{Decision Task Details}
\label{app:navigation}
\paragraph{Task Overview}
Decision-making tasks require an embodied agent to accomplish a given objective through a series of decisions. Each task environment is built using intricate 3D assets based on real-world references, ensuring accurate representation of structural and material characteristics. Figure~\ref{fig:perspective} illustrates the perceptual input at one specific state.

All decision tasks follow a unified operational protocol. For each task, the robot first queries its memory for the target's stored coordinates and navigates directly if found. If no prior data exists, it uses its six camera feeds to identify the target by visual comparison against a reference image, then executes a systematic exploration pattern to locate it. Throughout all missions, the robot must: (1) maintain a minimum standoff distance of 10 meters from all obstacles; (2) remain within predefined operational boundaries at all times; and (3) document all encountered artificial structures and special objects while excluding marine life from all reports.

\paragraph{Locate the Mining Robot}
Locate and approach the mining robot in an abandoned subsea research zone with variable terrain, low visibility, and artificial structures. The robot searches for the target by its distinct shape, structure, and color.

\paragraph{Inspect the Oil Pipeline}
Inspect the abandoned subsea oil pipeline network in a central zone. Pipelines may be partially buried and can serve as navigation references. The robot looks for linear structures and surface features matching the pipeline profile.

\paragraph{Locate Oil Drums}
Search for oil drums or barrels in a low-visibility environment, where they may be partially buried or scattered within sediment. The robot seeks out cylindrical objects and any visible markings indicative of oil drums.

\paragraph{Search for a Sunken Ship}
Detect sunken shipwrecks, which may be partially buried or obscured by underwater obstacles. The robot searches for large structural features and surface details characteristic of a shipwreck.

\paragraph{Locate the Electrical Box}
Find underwater electrical boxes, which are often partially buried in sediment and have distinctive structural features. The robot looks for specific shapes, structural traits, and identifiable markings.

\paragraph{Inspect the Wind Turbine}
Examine underwater wind power station structures, which are large installations with multiple pillars and mechanical components. The robot seeks out major structural and mechanical elements typical of wind turbines.

\paragraph{Search for the Aircraft}
Track down underwater aircraft wreckage, which may be complex and dispersed across different areas of the seafloor. The robot searches for key structural features and surface details indicative of aircraft debris.

\paragraph{Docking}
Approach an underwater landing platform marked with a distinctive "H" symbol, which serves as a reliable navigation reference. The robot looks for the "H" marking and the platform's regular structure.

\begin{figure*}[ht]
    \centering
    \includegraphics[width=0.95\textwidth]{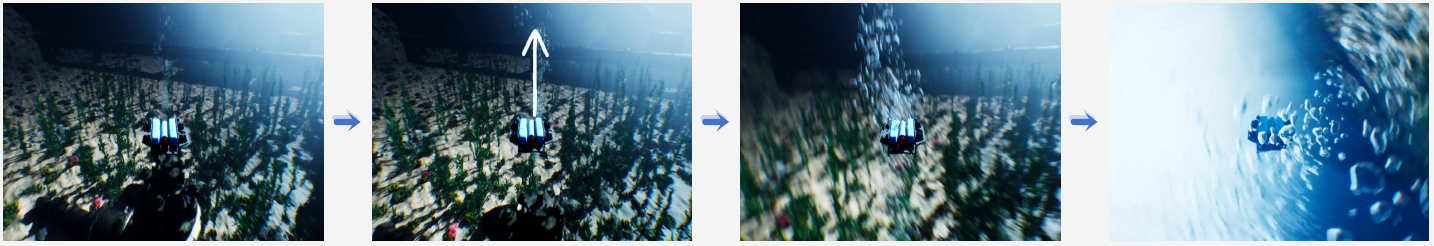}
    \caption{Examples of water flow interaction simulated using Unreal Engine plugins. An upward water flow is configured in the environment, exerting influence on an AUV as it passes through the affected region.}
    \label{fig:vid_frames}
\end{figure*}

\begin{table*}[t]
\centering
\small
\scalebox{0.96}{
\begin{tabular}{@{}lccccc@{}}
\toprule
\multirow{2}{*}{Model} & \multicolumn{5}{c}{Error Type (\%)} \\
\cmidrule(lr){2-6}
 & Missing All & Spurious All & No Overlap & \makecell{Partially Correct \\(Subset)} & \makecell{Partially Correct \\(Overlap)} \\
\midrule
\multicolumn{6}{c}{\textbf{Vision}} \\
\midrule
GPT-4o-mini & 94.85 & 0.00 & 3.68 & 0.74 & 0.74 \\
Gemini-2.5-Flash & 0.00 & 9.38 & 53.13 & 0.78 & 36.72 \\
Qwen2.5-VL-7B & 19.79 & 4.17 & 62.50 & 10.42 & 3.13 \\
Minicpm-4.5 & 2.97 & 9.90 & 53.47 & 7.92 & 25.74 \\
GPT-5.2 & 38.46 & 1.92 & 35.58 & 5.77 & 18.27 \\
Gemini-3-Pro & 14.13 & 7.61 & 18.48 & 5.43 & 54.35 \\
Human & 61.53 & 0.00 & 0.00 & 15.38 & 23.08 \\
\midrule
\multicolumn{6}{c}{\textbf{Vision+Sonar}} \\
\midrule
GPT-4o-mini & 86.09 & 0.00 & 7.52 & 1.88 & 4.51 \\
Gemini-2.5-Flash & 0.00 & 9.06 & 73.96 & 3.40 & 13.58 \\
Qwen2.5-VL-7B & 0.00 & 11.59 & 78.26 & 10.14 & 0.00 \\
Minicpm-4.5 & 0.81 & 9.76 & 71.14 & 7.72 & 10.57 \\
GPT-5.2 & 10.34 & 8.05 & 66.67 & 9.20 & 5.75 \\
Gemini-3-Pro & 2.30 & 11.49 & 28.74 & 8.05 & 49.43 \\
Human & 44.44 & 0.00 & 0.00 & 22.22 & 33.33 \\
\bottomrule
\end{tabular}}
\caption{Perception task error type metrics and results.}
\label{tab:error_analysis}
\end{table*}
\begin{table*}[t]
\centering
\small
\scalebox{0.95}{
\begin{tabular}{@{}lcccc@{}}
\toprule
Task & Target Not Found (\%) & Found but Blocked (\%) & Recognition Error (\%) & Memory Error (\%) \\
\midrule
1. Locate the robot (Shallow) & 67.65 & 17.65 & 8.82 & 5.88 \\
2. Locate the oil drums (Shallow) & 53.57 & 28.57 & 7.14 & 10.71 \\
3. Locate the electrical box (Shallow) & 53.85 & 23.08 & 7.69 & 15.38 \\
4. Search for a sunken ship (Shallow) & 60.87 & 4.35 & 13.04 & 21.74 \\
5. Search for the aircraft (Shallow) & 54.17 & 8.33 & 8.33 & 29.17 \\
6. Inspect the oil pipe (Shallow) & 50.00 & 26.67 & 23.33 & 0.00 \\
7. Inspect the wind turbine (Shallow) & 28.57 & 42.86 & 21.43 & 7.14 \\
8. Docking (Shallow) & 36.36 & 45.45 & 3.03 & 15.15 \\
9. Locate oil drums (Deep) & 75.00 & 5.56 & 13.89 & 5.56 \\
10. Search for a sunken ship (Deep) & 58.82 & 8.82 & 29.41 & 2.94 \\
11. Inspect the oil pipe (Deep) & 54.55 & 21.21 & 9.09 & 15.15 \\
12. Inspect the wind turbine (Deep) & 68.57 & 11.43 & 20.00 & 0.00 \\
\bottomrule
\end{tabular}
}
\caption{Decision task error type metrics and results.}
\label{tab:decision_error_analysis}
\end{table*}

\section{Error Type Metrics and Analysis}
\label{app:error}
As shown in the Table~\ref{tab:error_analysis}, we add auxiliary metrics for perception tasks to better characterize the types of errors made by the models. Specifically, we report the following error categories: (1) Missing All, where the ground-truth contains objects but the model fails to identify any of them; (2) Spurious All, where the ground-truth contains no objects but the model incorrectly reports one or more objects; (3) No Overlap, where the ground-truth contains objects and the model reports objects but there is no overlap between them; (4) Partially Correct (Subset), where the predicted objects are a correct subset of the ground-truth objects; and (5) Partially Correct (Overlap), where there is partial overlap between predicted and ground-truth objects. We find that \textbf{adding sonar helps both models and humans determine whether an object is present or not, but does not enable models to correctly identify what the object is}.
For decision tasks, we add auxiliary metrics (see Table~\ref{tab:decision_error_analysis}) to improve interpretability, focusing on failure mode analysis. Specifically, we categorize failures into four types: (1) Target Not Found, where the agent fails to locate the target due to ineffective search strategy; (2) Found but Blocked, where the target is located but subsequent task execution is obstructed; (3) Recognition Error, involving incorrect identification of the target or environment; and (4) Memory Error, involving failures in maintaining or recalling task-relevant information. We further observe that \textbf{in scenarios where the object is successfully found and normal operation proceeds, the proportion of memory errors versus recognition errors varies significantly across different tasks. This variation depends on factors such as object size and memory length requirements}.

% \section{Comparison with Non-MLLM Baseline}
% As shown in Table~\ref{tab:decision_more_methods}, we include a non-MLLM embodied baseline for decision tasks, specifically an underwater spiral search strategy. This heuristic policy progressively expands its search area while integrating YOLO-based object detection for target recognition. The results demonstrate that this non-MLLM baseline does not surpass the current best-performing MLLM agent.

% \input{table/decision_more_methods}

\section{Utilization of Unreal Engine Plugins}
Our simulation environment is compatible with Unreal Engine plugins. As shown in Figure~\ref{fig:vid_frames}, we leverage several UE plugins to simulate water interaction effects.

% \input{table/prompt}
% \input{table/decision_prompt}

% \section{The Impact of Different Prompts}
% Due to the difficulty in finding a prompt that is suitable for all MLLMs, we test the impact of different prompts on the model. As shown in Table~\ref{tab:prompt} and ~\ref{tab:decision_prompt}, we find that the impact was relatively small. Prompt1 is the prompt used in the main experiment, and prompt2 is the best-performing prompt among the alternatives tested for GPT-4o-mini.

\section{Prompt for {\bench}}
Figures~\ref{fig:prompt_perception}--\ref{fig:prompt_perception_sonar_examples} show the perception prompts, while Figure~\ref{fig:prompt_navigation} details the navigation prompt.

\newpage

\begin{figure*}[t]
\centering
\begin{tcolorbox}[
    breakable,
    width=\textwidth,
    title=Prompt for Perception Tasks,
    colback=white,
    colframe=blue!50!black,
    coltitle=white,
    fonttitle=\bfseries,
    before skip=10pt,
    after skip=10pt
]
\columnseprule=0.5pt

\textcolor{green}{\textbf{[RGB Image]}}

You are an assistant that analyzes an image and checks which of the following options appear in it.

Options:\textcolor{purple}{\textbf{[Options]}}

Instructions:

- Carefully examine the image, even the corners.

- You can choose single or multiple options, if none of the options appear, just return an empty list.

- For multiple-choice questions, no points will be awarded for incomplete selections, over-selections, or incorrect selections.

- The output must be a valid list (only list, no explanation, no extra text).
\end{tcolorbox}
\caption{Prompt template for perception tasks without sonar data.}
\label{fig:prompt_perception}
\end{figure*}

\begin{figure*}[t]
\centering
\begin{tcolorbox}[
    breakable,
    width=\textwidth,
    title=Prompt for Perception Tasks (Add Sonar),
    colback=white,
    colframe=blue!50!black,
    coltitle=white,
    fonttitle=\bfseries,
    before skip=10pt,
    after skip=10pt
]
\columnseprule=0.5pt

\textcolor{green}{\textbf{[Sonar Image]}}

This sonar image can be used as a reference to assist in identifying the next color image.\\

\textcolor{green}{\textbf{[RGB Image]}}

You are an assistant that analyzes an image and checks which of the following options appear in it. Before that, I have already provided you a sonar image to help you choose the correct one.

Options: \textcolor{purple}{\textbf{[Options]}}

Instructions:

- Only when you find it difficult to recognize the color image, I suggest you refer to the previous sonar image.

- Carefully examine the image, even the corners.

- You can choose single or multiple options, if none of the options appear, just return an empty list.

- For multiple-choice questions, no points will be awarded for incomplete selections, over-selections, or incorrect selections.

- The output must be a valid list (only list, no explanation, no extra text).
\end{tcolorbox}
\caption{Prompt template for perception tasks with sonar reference.}
\label{fig:prompt_perception_sonar}
\end{figure*}

\begin{figure*}[t]
\centering
\begin{tcolorbox}[
    width=\textwidth,
    title=Prompt for Perception Tasks (Add Sonar and Examples),
    colback=white,
    colframe=blue!50!black,
    coltitle=white,
    fonttitle=\bfseries,
    before skip=10pt,
    after skip=10pt
]
\columnseprule=0.5pt

\textcolor{green}{\textbf{[Object A Sonar Image]}}

This sonar image example is \textcolor{purple}{\textbf{[Object A]}}.\\

\textcolor{green}{\textbf{[Object B Sonar Image]}}

This sonar image example is \textcolor{purple}{\textbf{[Object B]}}.

\begin{center}
\textcolor{red}{\textbf{...}}
\end{center}

\textcolor{green}{\textbf{[Sonar Image]}}

This sonar image can be used as a reference to assist in identifying the next color image.\\

\textcolor{green}{\textbf{[RGB Image]}}

You are an assistant that analyzes an image and checks which of the following options appear in it. Before that, I have already provide you a sonar image to help you choose the correct one.

Options: \textcolor{purple}{\textbf{[Options]}}

Instructions:

- Only when you find it difficult to recognize the color image, I suggest you refer to the previous sonar image.

- Carefully examine the image, even the corners.

- You can choose single or multiple options, if none of the options appear, just return an empty list.

- For multiple-choice questions, no points will be awarded for incomplete selections, over-selections, or incorrect selections.

- The output must be a valid list (only list, no explanation, no extra text).
\end{tcolorbox}
\caption{Prompt template for perception tasks with sonar examples.}
\label{fig:prompt_perception_sonar_examples}
\end{figure*}

\begin{figure*}[t]
\centering
\begin{tcolorbox}[
    width=\textwidth,
    title=Prompt for Navigation Tasks,
    colback=white,
    colframe=blue!50!black,
    coltitle=white,
    fonttitle=\bfseries,
    before skip=10pt,
    after skip=10pt
]
\columnseprule=0.5pt
You are an expert pilot for an Autonomous Underwater Vehicle (AUV), designated as the "Control Expert". Your mission is to navigate a complex underwater environment to complete specific tasks. You will receive data from six cameras and location sensors. Your decisions must be precise, safe, and strategic.

1. Tactical Briefing for the Area of Operations

Before the mission begins, you must internalize the following intelligence about the operational area. This context is vital for interpreting sensor data and forming a macro-level strategy.

\begin{center}
\textcolor{red}{\textbf{...}}
\end{center}

3. Mission Briefing and Sensor Data

Task Description: \textcolor{purple}{\textbf{[Task Description]}}

Target Object Name: \textcolor{purple}{\textbf{[Object Name]}}

Target Object Reference Image: \textcolor{green}{\textbf{[Object Image]}}

Target Object Description: \textcolor{purple}{\textbf{[Object Description]}}

\begin{center}
\textcolor{red}{\textbf{...}}
\end{center}

5. Survey Navigation Commands

Available Commands: ``ascend'', ``descend'', ``move left'', ``move right'', ``move forward'', ``move backward'', ``rotate left'', ``rotate right'', ``stop''.

Command Execution: You must only issue one command per turn from the list above.

\begin{center}
\textcolor{red}{\textbf{...}}
\end{center}

Remember:

Conduct comprehensive reconnaissance. Systematic coverage is the priority. Use efficient exploration patterns. Catalog all special objects. Maintain exploration momentum. Always use the required format. Ignore all marine life. Keep responses as one continuous line between markers.
\end{tcolorbox}
\caption{Prompt template for navigation tasks.}
\label{fig:prompt_navigation}
\end{figure*}

\end{document}